\documentclass{article}
\usepackage[utf8]{inputenc}

\newif\ifconference
\conferencefalse

\usepackage{txie}

\makeatletter
\AtBeginDocument{%
  \let\cref@genuine@label\label
  \g@addto@macro\@arrayparboxrestore{\let\label\cref@genuine@label}%
}
\makeatother

\ifconference
  \usepackage[main]{neurips_2026}
\else
  \usepackage{geometry}
  \usepackage{parskip}
  \makeatletter
  \def\thm@space@setup{%
    \thm@preskip=\parskip
    \thm@postskip=\parskip
  }
  \makeatother
\fi

\usepackage[T1]{fontenc}
\ifconference\else\usepackage{lmodern}\fi
\usepackage[scaled=.91]{helvet}
\usepackage{inconsolata}
\usepackage{fvextra}
\usepackage{upquote}
\usepackage{microtype}
\usepackage{bm}

\usepackage[section]{placeins}
\usepackage{needspace}


\allowdisplaybreaks
\ifconference\else\emergencystretch=2em\fi

\title{Self-Improving Long-Horizon Reasoning by Learning Summaries from Local Continuations}
\author{Haoqun Cao\\
  University of Wisconsin--Madison\\
  \texttt{hcao65@wisc.edu}
  \and
  Tengyang Xie\\
  University of Wisconsin--Madison\\
  \texttt{tx@cs.wisc.edu}}
\date{}

\definecolor{BaseColor}{HTML}{D55E00}
\definecolor{TrainedColor}{HTML}{0072B2}
\newsavebox{\exboxsave}
\newcommand{\exgap}{\par\vspace{2.5pt}}
\newcommand{\pmse}[1]{{\scriptsize$\,\pm\,$#1}}
\newcommand{\exsym}[1]{\makebox[\fontcharwd\font`0][c]{\ensuremath{#1}}}
\newcommand{\exomit}{{\normalfont\itshape\color{black!55}[\ldots]}}
\newcommand{\exhl}{\bgroup\markoverwith{\textcolor{TrainedColor!16}{\rule[-0.6ex]{1pt}{2.45ex}}}\ULon}
\newenvironment{exbox}[2][\linewidth]{%
  \begin{lrbox}{\exboxsave}\begin{minipage}[t]{\dimexpr#1-2\fboxsep-2\fboxrule\relax}%
  \setlength{\parindent}{0pt}\setlength{\parskip}{0pt}%
  {\sffamily\bfseries\scriptsize #2\par}\vspace{1.5pt}\hrule\vspace{3pt}%
  \ttfamily\scriptsize\raggedright\everypar{\hangindent=1em\hangafter=1\relax}}%
 {\par\end{minipage}\end{lrbox}\fbox{\usebox{\exboxsave}}}
\newsavebox{\promptframesave}
\newenvironment{promptframe}{%
  \par\addvspace{\smallskipamount}\noindent
  \begin{lrbox}{\promptframesave}\begin{minipage}{\dimexpr\linewidth-12pt-2\fboxrule\relax}%
  \setlength{\parindent}{0pt}\setlength{\parskip}{0pt}\ttfamily\small\raggedright}%
 {\par\end{minipage}\end{lrbox}{\setlength{\fboxsep}{6pt}\fbox{\usebox{\promptframesave}}}\par\addvspace{\smallskipamount}}

\begin{document}
\maketitle

\begin{abstract}
Reasoning models degrade as their context grows, so longer reasoning traces do not by themselves give longer effective horizons. One way to extend the horizon is to compress the reasoning history into a compact state from time to time and continue from the state. Existing methods train the state with the final-answer reward or distill it from a stronger model. Neither signal is tailored to measure the contribution of the state alone. We observe that a reasoning model that degrades over long trajectories still behaves reliably in short contexts. We aim to preserve this short-context behavior when compressing the reasoning history. Motivated by this, we train a separate summarizer with Delta-KL, a reward computed from the frozen reasoning model's token-level predictions on sampled continuations, and run inference by truncating the trace, summarizing it, and continuing from the summary (TruSuCo). Training uses a few thousand questions with 4k-token reasoning prefixes and does not need answers or external teachers. Under a fixed 32k-token budget, the trained summarizer improves on vanilla reasoning and on an untrained summarizer in all nine settings across mathematical benchmarks and Reasoning Gym puzzles, by up to 7.7 points and 12 points, respectively. The improvement over vanilla reasoning grows with the budget. At 128k tokens, our method outperforms it by 10 points on AIME 2024--2026. Analyses show that truncation prevents the collapse of vanilla reasoning and that the trained summaries retain more of the reasoning history.
\end{abstract}

\section{Introduction}
Scaling test-time computation has become an effective approach to improving LLM reasoning~\citep{snell2024scaling,muennighoff2025s1}. To benefit from longer reasoning traces, however, a model must sustain useful progress over many steps.  Although many models support long context windows, this capacity does not guarantee reliable reasoning throughout the supported range and performance can deteriorate as context grows~\citep{hsieh2024ruler,du2025context}. Long-horizon reasoning therefore requires keeping the context short while retaining the information that later steps need.

A growing line of work addresses this challenge through context management, organizing reasoning into segments and periodically reducing the context used for continuation~\citep{yan2025inftythink,yan2026inftythinkplus,wu2026reasoningcache,aghajohari2025markovian,kontonis2026memento,yang2026accordion}. Across these approaches, reasoning continues from a compact representation of the preceding history, such as generated summaries or a retained portion of the original trace. This allows the reasoning horizon to extend while controlling the growth of the active context. The quality of the state carried between segments directly affects subsequent reasoning. Common training recipes to improve the quality of those states obtain supervision from task outcomes~\citep{yan2026inftythinkplus,wu2026reasoningcache,aghajohari2025markovian,yang2026accordion} or external teachers~\citep{yan2025inftythink,kontonis2026memento}. Outcome-based RL requires ground-truth answers or task verifiers. Furthermore, its feedback depends on whether subsequent reasoning produces a correct answer, which does not isolate the contribution of the state itself and may cause credit assignment issues, as explicitly discussed by \citet{wu2026reasoningcache}. {Other approaches learn how to construct states using summaries written by stronger models or training targets revised by external judges}~\citep{yan2025inftythink,kontonis2026memento}. These considerations motivate us to seek a dedicated learning signal for improving state quality without relying on final answer or a teacher.

Our choice of signal is guided by two considerations. First, \citet{yuan2026markovstates} show that estimating a compact Markov state improves sample efficiency by avoiding the potentially exponential coverage induced by the whole action history. A summary should therefore be approximately Markovian: conditioned on the summary, the subsequent reasoning of the solver (the model that performs the reasoning) should be nearly independent of the discarded history. Second, we also notice that the solver's reasoning remains reliable in short contexts even when it deteriorates over long trajectories, so its local behavior with short context can serve as a teacher. Together, we seek to replace a reasoning history with a compact state which approximately preserve the solver's continuation distribution at short context. This forms the principle of our supervision signal.

We introduce \textbf{TruSuCo} (Truncate, Summarize, Continue), an inference procedure that periodically replaces the solver's reasoning history with a compact summary state and continues generation from that state. We train the summarizer (a separate model that writes the summary) with \textbf{Delta-KL}, which evaluates candidate states through changes in the frozen solver's local continuation predictions. We first validate this approach in a controlled synthetic experiment consisting of independent arithmetic problems. Despite the simplicity of the individual problems, vanilla reasoning rapidly stops making progress as the context grows. The trained summarizer mitigates this collapse, allowing the solver to sustain progress while maintaining accuracy close to its initial short-context behavior.

We then evaluate on mathematical reasoning benchmarks and structured puzzles from Reasoning Gym. Under a fixed 32k-token solver budget, our method improves on vanilla reasoning and untrained summarization across all reported settings, achieving the highest accuracy in seven of nine evaluation settings. The gains reach 7.7 percentage points over vanilla reasoning and 12 percentage points over the untrained summarizer. The improvement over vanilla reasoning grows with the budget. On AIME 2024--2026, it reaches 10 percentage points at a 128k-token reasoning budget.

Our training is lightweight. We train the summarizer only using a few thousand questions, with 4k-token reasoning prefixes and precomputed continuation probes. For the 4B configuration, 80 optimization steps take approximately 5--6 hours on three H100 GPUs. \textbf{These gains demonstrate self-improvement only using the frozen solver's own continuation predictions, without ground-truth answers, task verifiers, outcome-based RL, or a stronger teacher.} Trajectory and summary analyses show that context reduction alleviates collapse, while training improves information retention and helps convert extended reasoning into correct solutions.

Our main contributions are as follows:
\begin{itemize}
    \item \textbf{An outcome-free signal for learning summaries.} We introduce Delta-KL, which scores candidate text summaries only using a frozen solver's own local continuation predictions and provides a dedicated signal for training the summarizer. This enables self-improvement without ground-truth answers, task verifiers, or a stronger teacher. Training is lightweight, only requiring a few thousand questions with short reasoning prefixes. Objective evaluations in Section~5.2 demonstrate that this signal improves summary quality, particularly information retention.

    \item \textbf{Benchmark gains and horizon generalization.} We combine an inference regime TruSuCo (Truncate, Summarize, Continue) with Delta-KL training to improve performance across math and puzzles. Under a fixed 32k-token solver budget, our method achieves the best performance in most evaluated settings. These gains generalize beyond the short contexts used during training, with performance continuing to improve as inference budgets scale to 128k tokens.
\end{itemize}

\section{Motivation}\label{sec:motivation}
Although a model can fail when its context is long, it actually reasons reliably when its context is short. This reminds us that improving long-horizon reasoning may not require external supervision. Below, we first formally state this intuition as a hypothesis and derive a procedure for long-horizon reasoning, and then verify both in a synthetic experiment.

\textbf{Accuracy can degrade in sufficiently long contexts.}
As discussed in the introduction, models become less reliable as their context grows {\citep{hsieh2024ruler,du2025context}}. Our working hypothesis is that a fixed language model $\pi_\theta$ can exhibit this degradation even when the additional context only consists of its own earlier output and is irrelevant to the current problem. Informally,
\begin{equation*}
    \pi_\theta(\cdot\,|\,C_{\mathrm{short}})\ \succ\ \pi_\theta(\cdot\,|\,C_{\mathrm{long}}), \qquad |C_{\mathrm{short}}| \ll |C_{\mathrm{long}}|,
\end{equation*}
where $\succ$ denotes ``behaves better than,'' $C_{\mathrm{short}}$ lies within the model's reliable short-context regime, and $C_{\mathrm{long}}$ extends it with substantial additional context.

\textbf{Improving long-horizon reasoning by compaction and imitation of short-context behavior.}
Notice that, under this hypothesis, improving long-horizon reasoning does not require changing $\theta$. The well-behaved version of $\pi_\theta$ is simply $\pi_\theta$ conditioned on a short context, therefore the model can serve as its own teacher. A long-horizon reasoning procedure should keep the model imitating $\pi_\theta$ conditioned on a short context.

This motivates the following procedure: \emph{(i) compact the context regularly, and (ii) at each compaction, require the compacted state to preserve the model's distribution over what comes next.} Note that both the compaction and the imitation must happen before $\pi_\theta$ degrades, so that the distribution being imitated is still a reliable one. Formally, we look for a state $S$ such that, whenever $\pi_\theta(\cdot\,|\,C)$ has not degraded,
\begin{equation}\label{eq:intuition}
    \pi_\theta(\cdot\,|\,S) \approx \pi_\theta(\cdot\,|\,C), \qquad |S| \ll |C|.
\end{equation}
Applied repeatedly, this procedure compacts the context after every stretch of reasoning and continues from $S$, so the context stays short while the model always operates in the regime in which it has not degraded. We first verify the hypothesis and the procedure in a synthetic experiment, and then describe the methodology in \cref{sec:methodology}.

\subsection{A Synthetic Verification}
\textbf{Setup.}
We illustrate this intuition with a synthetic task. A problem instance is a single prompt with 64 independent multiplication problems, each a three-digit number times a two-digit number. The model is asked to solve them in order, writing ``Answer $k = v$'' after each. The solver is DeepSeek-R1-Distill-Qwen-1.5B and is kept frozen. We use 24 problem instances and sample 16 runs per instance for each compared method, with a budget of 49{,}152 generated tokens per run. The prompt is given in Appendix~\ref{app:synthetic}.

\textbf{Scoring.}
We score the answer to every one of the 64 problems. For each problem we take the first answer the solver gives, so it has no chance to correct an answer later. A problem that is never answered within the budget is counted as wrong.

\textbf{Compared methods.}
We compare six ways of running the solver. \emph{Single} is the vanilla reasoning. 
The other five methods share the same regime. We cut the reasoning trace of the solver every 2{,}048 tokens, and the solver's context is then replaced by the original question and a state of at most 256 tokens. The solver then continues from this context. The five methods differ in how the state is produced. \emph{Copy} keeps the last 256 tokens of the trace verbatim. \emph{Delethink} keeps the first 100 tokens of the first chunk and the last 156 tokens of the trace. \emph{Base} and \emph{Trained} keep the last 128 tokens of the trace verbatim and place a summary of at most 128 tokens before them. Base uses an untrained Qwen3-4B-Instruct as the summarizer. Trained uses the same summarizer after training it with the condition in \eqref{eq:intuition} on states collected from this synthetic task, with the frozen solver as the only source of supervision. The detailed training procedure is formally introduced in Section~\ref{sec:methodology}. {\emph{Oracle} replaces the summary with a statement written by a rule. The statement tells the solver which problems are done and which one it is working on.}  An example is given in Appendix~\ref{app:synthetic}.

\textbf{Results.}
\cref{fig:e0} shows the accuracy on the 64 problems, averaged in groups of eight. The dashed line is the accuracy of the answers the solver gives before the first cut. At that point its context is still clean, so it represents the accuracy of the solver in its normal working state (we call working accuracy). Without truncation, the solver moves away from this line as it proceeds and ends below 0.05 on the last problems. Every method that drops part of the context avoids this collapse and stays above 0.6 on all groups of problems. Among them, Trained is the closest to Oracle and to the working accuracy, and it stays at about 0.8. Over all 64 problems, the share of correct answers is 46.9\% for Single, 74.0\% for Copy, 70.7\% for Delethink, 72.9\% for Base, 82.2\% for Trained and 87.5\% for Oracle.

\textbf{Takeaways.}
First, the solver without truncation collapses as it goes on, even though later problems are unrelated to earlier ones. The length of the context harms its performance. Second, every method that drops part of the context helps, because dropping makes the context short again. Even copying the last 256 tokens of the trace is enough for the solver to carry on. Third, the trained summarizer is closer to the working accuracy than every other method except the oracle. Oracle uses a task-specific rule to tell the solver where it is in the list. It shows how close the solver can come to its working accuracy when given a clear state. The trained summarizer comes close to it after training. This is because the summary is chosen to imitate the behavior of the model in the regime where it works well, so more of that accuracy carries over to the rest of the list. \cref{fig:e0} (right) shows what the two summarizers write in one run. The untrained summarizer writes a simple and complete problem at the point where the reasoning stops. However, the sentence does not really look like a summary. It looks like a complete and isolated question to solve. Then the solver goes on to solve this question in isolation and thinks that the task has ended. The trained summarizer provides a more comprehensive summary and writes sentences that confirm the progress. More importantly, it actually induces the solver to continue with the rest of the problems, which exactly matches our intuition in \eqref{eq:intuition}. We also find that 90.0\% of the trained summaries state which problems are done, against 51.8\% of the untrained ones. This helps the solver to keep track of its progress and carry on. The two summarizers receive the same instruction, so the difference comes from training. Appendix~\ref{app:synthetic} gives the full example and further details.

\begin{figure}[!htbp]
    \centering
    \begin{minipage}[c]{0.535\linewidth}
        \centering
        \includegraphics[width=\linewidth]{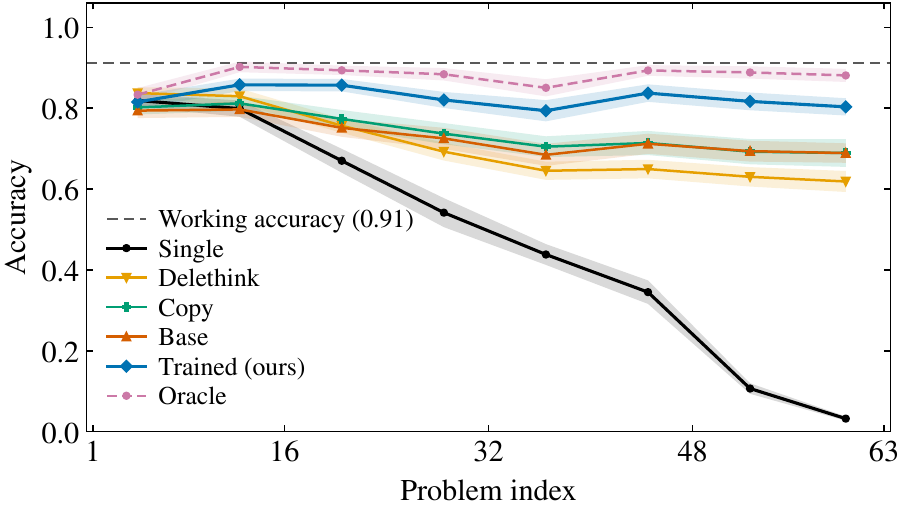}
    \end{minipage}\hfill
    \begin{minipage}[c]{0.45\linewidth}
        \begin{exbox}{The solver's reasoning at the first cut}
        \exomit{}\par
        Problem 13: 667 \exsym{\times} 34. Breaking this down: 667 \exsym{\times} 30 is 20,010 and 667 \exsym{\times} 4 is 2,668.\par
        \exomit{}\par
        600\exsym{\times}34 is 20,400 and 67\exsym{\times}34 is 2\par
        \end{exbox}\par\vspace{4pt}
        \begin{exbox}{\textcolor{BaseColor}{Base summary}}
        Okay, so let me recap where I am.\par
        \exgap
        Problem 13: 667 \exsym{\times} 34.\par
        \end{exbox}\par\vspace{1pt}
        {\scriptsize\sffamily\raggedright $\rightarrow$ The solver answers Problem 13 and stops: 13 of 64 right.\par}\vspace{4pt}
        \begin{exbox}{\textcolor{TrainedColor}{Trained summary}}
        Okay, so let me recap where I am.\par
        \exgap
        Okay, so \exhl{I've gone through Problems 1 through 12, breaking each multiplication down using distributive properties and step-by-step addition. All answers are verified via two methods where possible.}\par
        \exgap
        Problem 13: 667 \exsym{\times} 34.\par
        \end{exbox}\par\vspace{1pt}
        {\scriptsize\sffamily\raggedright $\rightarrow$ The solver goes on to Problem 14: 63 of 64 right.\par}
    \end{minipage}
    \caption{\textbf{Left:} Accuracy on the 64 independent multiplication problems, averaged in groups of eight. Each of the 24 instances is sampled 16 times. The last point covers Problems 57 to 63, and shaded bands show one standard error over the instances. The dashed line is the working accuracy, the accuracy of the answers given before the first cut, pooled over methods. Without truncation the solver declines steadily to almost zero. All truncation methods stay above 0.6, and Trained stays close to Oracle. \textbf{Right:} one run at its first cut, which falls in the middle of Problem 13.}
    \label{fig:e0}
\end{figure}
\section{Methodology}\label{sec:methodology}

We introduce the inference procedure TruSuCo (Truncate, Summarize, Continue) and the training objective Delta-KL. TruSuCo uses a frozen LLM solver $\pi_{\mathrm{so}}$ and a separate LLM $g$ as its summarizer. We only train the summarizer.

{Let $K$ and $k$ denote the lengths of the local reasoning window and the retained tail, respectively. We use $X$ for the question, $\oplus$ for concatenation, and $P_{\mathrm{so}}$ and $P_{\mathrm{su}}$ for the solver and summarizer instructions. The solver's chat template, $\mathrm{chat}_{\mathrm{so}}$, wraps its input in a single user message and ends at the start of the assistant turn, including \texttt{<think>} for a thinking model. We write a summarizer call as $g(X,\ldots)$, leaving its fixed instruction $P_{\mathrm{su}}$ and chat template implicit.}

\paragraph{Inference.}
{At round $t$, the solver continues from state $s_t$. The retained tail at the end of $s_t$ and the newly generated reasoning (from the solver) together form the local window $a_t$. We start with $s_1=\varnothing$, so the first window only contains new reasoning. Once $a_t$ reaches $K$ (e.g. 4096) tokens, we cut the trace, keep its last $k$ (e.g. 128) tokens as the new tail, and only summarize the first $K-k$ tokens of $a_t$. Each summary has a token budget (e.g. 512) well below the length of the reasoning it replaces. If the solver finishes or exhausts its total budget, we stop without updating the state.}

Notably, when the solver continues from the state $s_t$, the state is appended directly to the solver's open assistant turn, after the instruction and question formatted by its chat template. The solver reads the summaries and tail as its own ongoing reasoning and continues from there.

At the cut in round $t$, we generate a summary $m_t$ and form the next state $s_{t+1}$ in one of two ways. In \emph{append} mode, each window is summarized independently, and we concatenate all summaries produced so far. In \emph{roll} mode, the summarizer also reads the previous summary, and the old summary will not be kept. The first round starts with an empty previous summary, $m_0=\varnothing$. Writing ``latest tail'' for the last $k$ tokens of $a_t$, the two updates are:
\begin{align*}
\text{append:}\qquad m_t
  &\sim g\bigl(X,(a_t)_{1:K-k}\bigr),\\
s_{t+1}
  &=m_1\oplus\cdots\oplus m_t\oplus\text{latest tail},\\[3pt]
\text{roll:}\qquad m_t
  &\sim g\bigl(X,m_{t-1},(a_t)_{1:K-k}\bigr),\\
s_{t+1}
  &=m_t\oplus\text{latest tail},\\[3pt]
\text{next solver input}
  &=\mathrm{chat}_{\mathrm{so}}(P_{\mathrm{so}}\oplus X)\oplus s_{t+1}.
\end{align*}
Appendix~\ref{app:implementation} specifies the values of those parameters used in the experiments.
Appendix~\ref{app:main_prompts} gives the solver and summarizer prompts for the main experiments.

\paragraph{Training.}
During training, the state contains a single summary and the latest tail.
For each training question $X$, we collect the frozen solver's first $K$ reasoning tokens as the window $a$. As in inference, the summarizer reads $X$ and $a_{1:K-k}$ to produce $m$, and we form $s=m\oplus\text{latest tail}$. We sample $J$ continuation probes $y^{(j)}\sim\pi_{\mathrm{so}}(\cdot\mid Q\oplus a)$ from the full prefix once and reuse them throughout training, where $Q=\mathrm{chat}_{\mathrm{so}}(P_{\mathrm{so}}\oplus X)$ is the fixed solver prompt. To score a candidate, we teacher-force these same probe tokens under the full prefix and the candidate state:
\begingroup
\begin{equation}\label{eq:dkl}
\begin{aligned}
\Delta_{j,i}(s)
 &=\log\pi_{\mathrm{so}}\bigl(y_i^{(j)}\mid Q\oplus s\oplus y_{<i}^{(j)}\bigr)\\
 &\quad-\log\pi_{\mathrm{so}}\bigl(y_i^{(j)}\mid Q\oplus a\oplus y_{<i}^{(j)}\bigr).
\end{aligned}
\end{equation}
\endgroup
 These log-probability differences measure how compression changes the solver's continuation distributions. 
Uniformly averaging \eqref{eq:dkl} estimates the negative KL divergence between the full-prefix continuation distribution and the state-conditioned one. In practice, this objective strongly favors copying the tail $s^{\mathrm{tail}}$ (the final tokens of $a$, with the same budget as one summary plus the retained tail), which can lose information needed later.
We therefore reweight the KL objective to discourage this shortcut. This reweighting acts as a practical regularizer, giving more weight to probe positions where the full prefix predicts better than $s^{\mathrm{tail}}$.
The weight of position $i$ in probe $j$ and the total weight of the probe are
\begin{equation}\label{eq:weights}
 w_{j,i}=\max\bigl(0,-\Delta_{j,i}(s^{\mathrm{tail}})\bigr),
 \qquad Z_j=\sum_{i\in\mathcal I_j}w_{j,i},
\end{equation}
where $\mathcal I_j$ is the set of scored positions in probe $j$. We then take the weighted average of the differences in \eqref{eq:dkl}. Positive differences are first clipped at zero, which prevents gains on some tokens from offsetting losses on others. Each probe is normalized by its total weight, and the average runs over the probes with positive total weight, $\mathcal J_+=\{j:Z_j>0\}$:
\begin{equation}\label{eq:rdkl}
 R_{\Delta\mathrm{KL}}(s)=\frac{1}{|\mathcal J_+|}\sum_{j\in\mathcal J_+}
  \frac{\sum_{i\in\mathcal I_j}w_{j,i}\min\bigl(\Delta_{j,i}(s),0\bigr)}{Z_j}.
\end{equation}
We also apply a length penalty. Let $\operatorname{cap}(m)$ be the event that the summary $m$ reaches its generation limit. In this case we subtract a constant $\lambda$, which encourages the summarizer to finish within its budget:
\begin{equation}\label{eq:final_reward}
 r(s)=R_{\Delta\mathrm{KL}}(s)-\lambda\,\mathbf1\{\operatorname{cap}(m)\}.
\end{equation}
We call $r(s)$ the Delta-KL reward. Appendix~\ref{app:implementation} specifies the valid probe positions, the treatment of zero weights, and when the length penalty is applied.

At each step, we draw $G$ fresh summaries per question and update $g$ with GRPO~\citep{shao2024deepseekmath} using group-normalized rewards. The solver and stored probes stay fixed (Algorithm~\ref{alg:delta_kl}).

\begin{algorithm}[!htbp]
\caption{Delta-KL training of the summarizer}
\label{alg:delta_kl}
\begingroup
\small
\algtext*{EndFor}
\begin{algorithmic}[1]
\Require $\pi_{\mathrm{so}}$ (frozen), $g$ (trainable), questions $\mathcal D$
\Statex \textbf{Precompute once}
\For{$X\in\mathcal D$}
    \State $a\sim\pi_{\mathrm{so}}(\cdot\mid Q)$ \Comment{$K$ tokens, unfinished only}
    \State $y^{(j)}\sim\pi_{\mathrm{so}}(\cdot\mid Q\oplus a),\quad j=1,\ldots,J$
    \State Cache $(X,a,\{y^{(j)}\})$ and full/suffix log-probabilities
\EndFor
\Statex \textbf{Training}
\For{each training step}
    \For{$(X,a,\{y^{(j)}\})$ in a sampled minibatch $\mathcal B$}
        \State $m_X^{(h)}\sim g(X,a_{1:K-k}),\quad h=1,\ldots,G$
        \State $s_X^{(h)}\gets m_X^{(h)}\oplus a_{K-k+1:K},\quad r_{X,h}\gets r(s_X^{(h)})$ \Comment{\eqref{eq:final_reward}}
    \EndFor
    \State $g\gets\operatorname{GRPO}\bigl(g,\mathcal B,\{m_X^{(h)},r_{X,h}\}_{X,h}\bigr)$ \Comment{group-normalized rewards}
\EndFor
\end{algorithmic}
\endgroup
\end{algorithm}

\section{Experimental Setup}
We evaluate the training method Delta-KL and the inference procedure TruSuCo on two families of long-horizon reasoning tasks: general math, and three puzzles from Reasoning Gym~\citep{stojanovski2025reasoninggym}, namely Zebra, Sokoban and Futoshiki. Appendix~\ref{app:rg_tasks} describes the three puzzles. Below we introduce the detailed setup.

\paragraph{Models.}
We use two solvers: DeepSeek-R1-Distill-Qwen-1.5B and Qwen3-4B-Instruct-2507. Both have a relatively short effective context, which leaves room for improving their long-horizon reasoning. For the summarizer we use instruction-following models, since the summarizer has to read another model's reasoning trace and follow a fixed instruction, and these models retain enough reasoning ability in their base model to do so. We use Qwen3-1.7B (non-thinking mode) and Qwen3-4B-Instruct-2507.

\paragraph{Training data.}
For general math, the training data is built from DeepScaleR~\citep{deepscaler}. We take a subset of its questions and let the solver reason on each one for $4096$ tokens. We keep the questions on which the solver has not finished within these tokens, and for each of them sample $J$ probes from the solver conditioned on the full $4096$ tokens. After filtering, the R1-1.5B and Qwen3-4B-Instruct datasets contain $2{,}514$ and $2{,}490$ questions, respectively. For Reasoning Gym, we generate 1{,}200 puzzles at each of two difficulty levels for every task and collect prefixes and probes in the same way. We train a separate summarizer for each task. Ground-truth answers are neither provided to the summarizer nor used to compute its reward.

\paragraph{Evaluation data.}
For mathematics, we evaluate on 220 problems: 90 from AIME 2024--2026, 100 from BeyondAIME~\citep{bytedanceseed2025beyondaime}, and 30 from HMMT February 2025. For Reasoning Gym, we use the procedural generators to construct 192 test problems per task, with 96 at each of two difficulty levels. For Zebra, the harder of the two test levels is not used in training. Appendix~\ref{app:rg_tasks} gives the difficulty levels, the generation parameters and seeds.

\paragraph{Baselines.}
We compare inference regimes for the same frozen solver. For a fair comparison, we exclude variants that use ground-truth answers, outcome rewards, or stronger teachers for additional training. \emph{Single} generates without interruption until the budget is exhausted (i.e. vanilla reasoning). \emph{Base} is TruSuCo with the untrained summarizer, and \emph{Trained} is TruSuCo with the summarizer trained by Delta-KL. \emph{RC} is the Reasoning Cache regime~\citep{wu2026reasoningcache}: the solver makes successive attempts of 16k tokens, and after each attempt writes a cache of at most 2{,}048 tokens with Reasoning Cache's own prompt, from which the next attempt starts. The number of attempts is the budget divided by 16k. \emph{Delethink} is the inference regime of the Markovian Thinker~\citep{aghajohari2025markovian}: generation proceeds in chunks of $C$ tokens, and after each chunk the context is reset to the question, the first 100 tokens of the first chunk, and the last $M = C/2$ tokens of the previous chunk, with $C = 8$k for the 4B solver and $C = 4$k for R1-Distill-1.5B. Appendix~\ref{app:eval_protocol} gives the remaining details of both.

\paragraph{Training protocol.}
We train the summarizer with GRPO in verl and keep the solver frozen. Each training set contains about 2{,}000 to 2{,}500 examples. Each step uses a batch of 32 examples and samples eight candidate summaries for each, and every candidate is scored on the eight stored probes of its example, which are up to 1{,}024 tokens long. We report results from the checkpoint after 80 steps. This amounts to 2{,}560 examples, about one pass over the training set for the math experiments. Appendix~\ref{app:implementation} lists all training settings.

\paragraph{Evaluation protocol.}
At inference, TruSuCo uses a reasoning window of $K=4096$ tokens, a retained tail of $k=128$ tokens and a cap of 512 tokens per summary, the same as in training. Appendix~\ref{app:implementation} gives the remaining inference settings. We evaluate every method under the same solver budget, which only counts the tokens the solver generates, accumulated over chunks. A run terminates naturally if the solver closes its reasoning block or emits \texttt{[EOS]} within the budget. Otherwise it is forced to answer. We add a fixed sentence saying that the thinking budget has run out and close the reasoning block. The solver then has up to 1,024 extra tokens to answer. Appendix~\ref{app:eval_protocol} gives the remaining details.

\section{Results and Analysis}\label{sec:results}

\cref{tab:main_results} reports the main results. \textbf{Under a 32k solver-token budget, Trained improves over Single and over Base in all nine settings, and it is the best method in seven of them.} The gains reach 7.7 points over Single and 12 points over Base. In the other two settings, BeyondAIME with the 4B solver and HMMT with the R1 solver, Trained is within one point of the best method. It is also the only method that improves over Single in every setting. \textbf{These gains are obtained without outcome supervision or external teachers.} The only training signal is the solver's own behavior on short contexts, so we view the method as a form of self-improvement.

The gains hold in every setting, however their size varies with the task. Over Base, the gain is at most about 2 points on mathematics and between 3 and 12 points on Reasoning Gym. On mathematics the untrained summarizer is already useful and improves over Single in all six settings, so the room for training is smaller. On Sokoban and Futoshiki the untrained summarizer falls below Single, by 3.5 and 11 points. Training reverses this, and Trained is the best method on both puzzles. The two baselines RC and Delethink are less consistent. RC is ahead of Trained on BeyondAIME with the 4B solver and on HMMT with the R1 solver, by less than one point, and it falls 2.8 points below Single on Futoshiki. Delethink is 0.1 point above Trained in one setting and below it in the other eight, and it falls 1.8 points below Single on Sokoban. Since the size of the gain differs across tasks, we next examine where it comes from.

{\textbf{Does truncation alone explain the gains, or does summary content also matter?} Removing history can interrupt unproductive generation, but it can also discard useful intermediate results. We compare reasoning trajectories under the different inference regimes, then ask \textbf{what does training change in the summaries?} We compare Base and Trained on information retention and other summary properties, and examine which properties the reward favors.}

\begin{table*}[!htbp]
    \centering
    \caption{
        Main results on mathematical reasoning and Reasoning Gym under a
        32k solver-token budget.
        Accuracy is reported in percent, as the mean over 32 runs per problem (avg@32),
        followed by one standard error over problems.
        $\Delta_{\mathrm{Single}}$ and $\Delta_{\mathrm{Base}}$ denote
        improvements of Trained over Single and Base, respectively,
        in percentage points. They are computed before rounding, and their
        standard errors are computed from the per-problem differences.
        Boldface indicates the highest mean accuracy in each row.
    }
    \label{tab:main_results}
    \small
    \setlength{\tabcolsep}{3pt}
    \renewcommand{\arraystretch}{1.15}

    \begin{tabular*}{\textwidth}{
        @{\extracolsep{\fill}}lrrrrrrr@{}
    }
        \toprule
        Dataset / Task
        & Single
        & Base
        & RC
        & Delethink
        & Trained
        & $\Delta_{\mathrm{Single}}$
        & $\Delta_{\mathrm{Base}}$ \\
        \midrule

        \multicolumn{8}{@{}l}{
            \textbf{(a) Mathematics: 1.7B summarizer + R1-Distill-1.5B solver}
        } \\
        \addlinespace[3pt]

        AIME 24--26 (90)
        & 24.0\pmse{3.5} & 27.3\pmse{3.6} & 26.2\pmse{3.7} & 26.0\pmse{3.6} & \textbf{28.5}\pmse{3.7}
        & $+4.5$\pmse{1.0}
        & $+1.2$\pmse{0.8} \\

        BeyondAIME (100)
        & 9.8\pmse{1.9} & 10.4\pmse{2.1} & 11.4\pmse{2.2} & 10.3\pmse{2.1} & \textbf{11.9}\pmse{2.2}
        & $+2.2$\pmse{0.8}
        & $+1.5$\pmse{0.7} \\

        HMMT Feb 2025 (30)
        & 13.6\pmse{5.1} & 13.8\pmse{5.2} & \textbf{14.9}\pmse{5.3} & 14.4\pmse{5.2} & 14.3\pmse{5.1}
        & $+0.6$\pmse{1.2}
        & $+0.5$\pmse{1.5} \\

        \emph{All (220)}
        & 16.1\pmse{1.9} & 17.8\pmse{2.0} & 17.9\pmse{2.0} & 17.3\pmse{1.9} & \textbf{19.0}\pmse{2.0}
        & $+2.9$\pmse{0.6}
        & $+1.3$\pmse{0.5} \\

        \midrule

        \multicolumn{8}{@{}l}{
            \textbf{(b) Mathematics: 4B summarizer + Qwen3-4B-Instruct solver}
        } \\
        \addlinespace[3pt]

        AIME 24--26 (90)
        & 53.4\pmse{4.1} & 59.6\pmse{4.2} & 60.2\pmse{4.2} & 58.6\pmse{4.1} & \textbf{61.1}\pmse{4.1}
        & $+7.7$\pmse{1.4}
        & $+1.6$\pmse{0.8} \\

        BeyondAIME (100)
        & 31.8\pmse{3.7} & 35.2\pmse{3.9} & \textbf{36.8}\pmse{3.9} & 34.9\pmse{3.8} & 36.6\pmse{3.9}
        & $+4.8$\pmse{1.1}
        & $+1.3$\pmse{0.8} \\

        HMMT Feb 2025 (30)
        & 29.9\pmse{7.3} & 34.1\pmse{7.7} & 33.9\pmse{7.7} & 34.9\pmse{7.4} & \textbf{35.7}\pmse{7.8}
        & $+5.8$\pmse{2.0}
        & $+1.7$\pmse{1.2} \\

        \emph{All (220)}
        & 40.4\pmse{2.7} & 45.0\pmse{2.8} & 45.9\pmse{2.8} & 44.6\pmse{2.7} & \textbf{46.5}\pmse{2.8}
        & $+6.1$\pmse{0.8}
        & $+1.5$\pmse{0.5} \\

        \midrule

        \multicolumn{8}{@{}l}{
            \textbf{(c) Reasoning Gym: 4B summarizer + Qwen3-4B-Instruct solver}
        } \\
        \addlinespace[3pt]

        Zebra
        & 48.1\pmse{2.5} & 51.5\pmse{2.4} & 53.8\pmse{2.5} & 52.2\pmse{2.3} & \textbf{54.6}\pmse{2.4}
        & $+6.6$\pmse{1.0}
        & $+3.2$\pmse{0.8} \\

        Sokoban
        & 37.5\pmse{2.5} & 34.0\pmse{2.6} & 39.9\pmse{2.6} & 35.7\pmse{2.5} & \textbf{41.6}\pmse{2.5}
        & $+4.1$\pmse{0.7}
        & $+7.6$\pmse{0.8} \\

        Futoshiki
        & 45.2\pmse{2.4} & 34.2\pmse{2.5} & 42.4\pmse{2.5} & 45.9\pmse{2.6} & \textbf{46.1}\pmse{2.5}
        & $+0.9$\pmse{0.9}
        & $+12.0$\pmse{1.0} \\

        \bottomrule
    \end{tabular*}
\end{table*}

\FloatBarrier

\subsection{Reasoning Trajectories across Inference Regimes}

{We compare reasoning trajectories on the same problems to examine how inference regimes affect answer timing, repetitive generation, and final accuracy. As a control for context reduction, we include \emph{Copy}, which replaces each summary state with a verbatim copy of the most recent reasoning tokens under the same state-token budget. This comparison helps distinguish the effects of truncating the context from those of constructing a summary.}

\begin{figure*}[!htbp]
    \centering
    \includegraphics[width=0.85\textwidth]{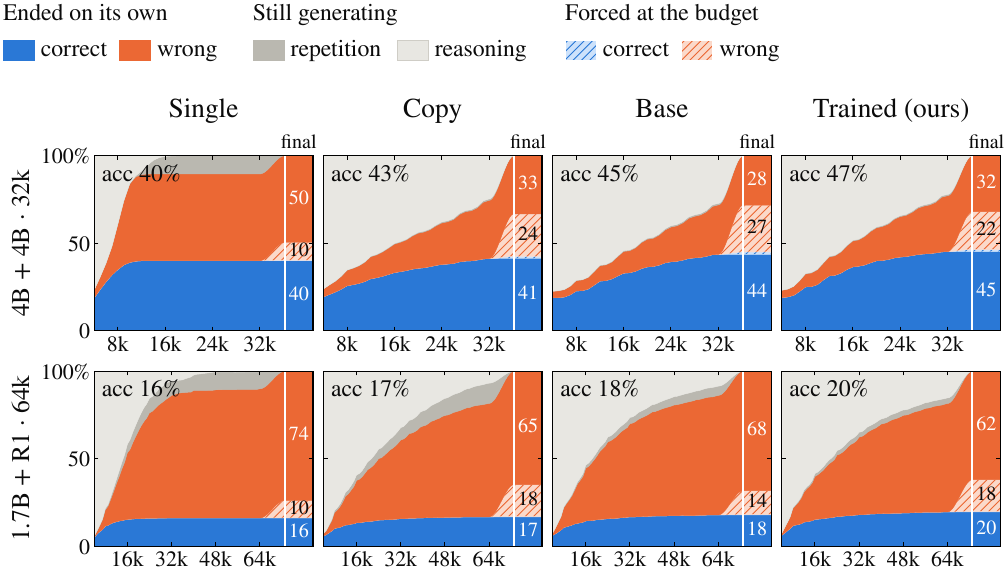}
    \caption{
        {Reasoning trajectories on the mathematics problems.
        Stacked areas show the fractions of runs that have ended on
        their own with correct or incorrect answers, or are still
        generating with or without detected repetitive degeneration. Hatched regions
        indicate answers obtained by forced finalization at the budget.
        The final bar in each panel shows the outcome proportions after
        forced finalization. Each row corresponds to a model
        configuration and each column to an inference regime.}
    }
    \label{fig:math_states}
\end{figure*}

\paragraph{Mathematics.}

{Single typically produces answers early, and its accuracy soon
plateaus (\cref{fig:math_states}). Some runs that remain active enter
\emph{repetitive degeneration}: sustained loops of repeated text, which
we detect programmatically from the generated output. Context truncation
substantially reduces this behavior{. With the 4B solver}, it is nearly
absent in Copy, Base, and Trained. These methods produce answers later, but the share of runs that finish correctly keeps rising. Copy, Base, and Trained shorten the context in the same way and only differ in what the state contains. Their final accuracies still differ, and Trained is above Copy in both configurations. So the accuracy also depends on the content of the state.}

\begin{figure*}[!htbp]
    \centering
    \includegraphics[width=0.85\textwidth]{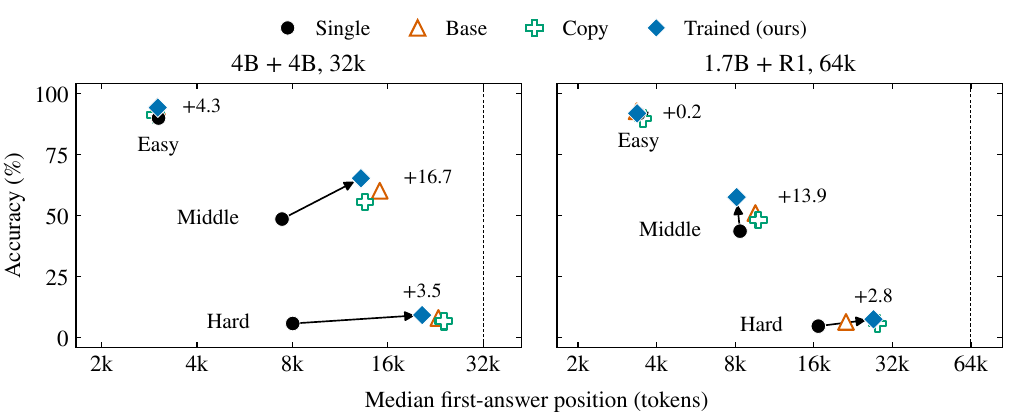}
    \caption{
        {Answer timing and final accuracy across problem difficulties.
        Each panel shows a model configuration. Points give the median
        first-answer position and final accuracy for
         Single, Base, Copy, and Trained
        within each difficulty group, defined by Single's
        accuracy on an independent evaluation run. Arrows connect Single
        to Trained{. The numbers give} accuracy gains in percentage points.
        The horizontal axis is logarithmic, and the dashed vertical line
        marks the solver-token budget.}
    }
    \label{fig:math_difficulty}
\end{figure*}

{The benefit of extended reasoning depends on problem difficulty
(\cref{fig:math_difficulty}). In both model configurations,
the Single-to-Trained arrows show the largest accuracy gains on problems
of intermediate difficulty. Easy problems
are often already solved by Single with substantially fewer tokens,
whereas accuracy on hard problems remains low even when answers are
produced much later. In the intermediate group, Trained also improves
accuracy over Base and Copy while producing its first answer at an earlier median
token position.}

\FloatBarrier

\begin{figure*}[!htbp]
    \centering
    \includegraphics[width=0.85\textwidth]{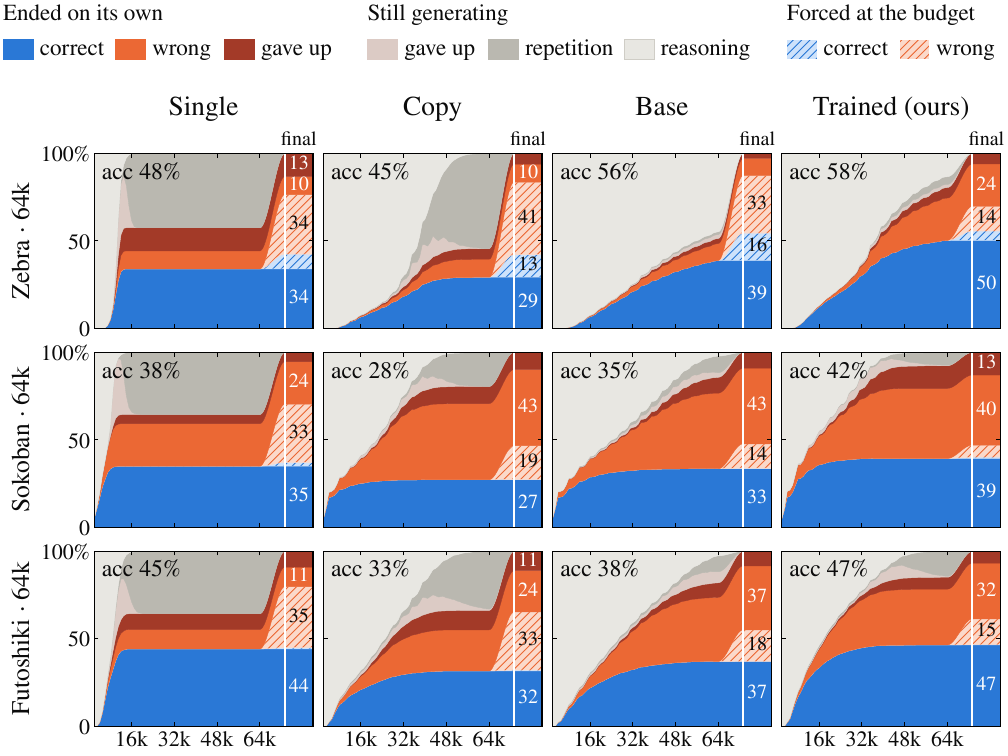}
    \caption{
        {Reasoning trajectories on Reasoning Gym under a 64k
        solver-token budget. Rows correspond to tasks and columns to
        inference regimes. Stacked areas distinguish correct and
        incorrect normal termination, give-up behavior, repetition, and
        continued generation. Hatched regions mark answers obtained by
        forced finalization at the budget{. The bars on the right} show
        the final outcome proportions.}
    }
    \label{fig:rg_trajectories}
\end{figure*}

\paragraph{Reasoning Gym.}
{Reasoning Gym reveals an explicit sign of long-context degradation
(\cref{fig:rg_trajectories}): Single often states that it is giving up, an event we detect
programmatically. Most such trajectories subsequently enter repetitive
generation, consistent with the behavior observed in mathematics.
Copy can delay this transition, but Trained
exhibits substantially fewer give-up events and less repetition than
Single and Copy.}

{Trained also yields a larger fraction of normally terminated,
correct runs than Single, Copy, and Base. Relative to Base, this fraction
increases from 39\% to 50\% on Zebra, 33\% to 39\% on Sokoban, and 37\% to
47\% on Futoshiki. These gains suggest that learned summaries may better preserve the
information needed for continuation. The next subsection examines this
explanation through direct comparisons of summary content.}

\FloatBarrier

\subsection{Summarizer Behavior Before and After Training}\label{sec:summary_behavior}
The results above show that the trained summarizer leads to better reasoning trajectories. So what does the trained summarizer actually learn to do differently? One observation is that, although our training objective never directly optimizes summary quality, it  substantially improves the quality of the resulting summaries. \textbf{This suggests that local continuation preservation serves as an effective proxy for summary quality}.

\paragraph{Summary Quality Comparison}
We compare Base and Trained on matched reasoning contexts using four
metrics. \textbf{Source-item recall} is the fraction of extracted source
items retained in a summary. The \textbf{bad-item rate} is the fraction of
extracted summary items absent from or inconsistent with the source.
Both use fixed task-specific extractors and exclude the retained tail.
We also report two more metrics. The \textbf{cap-hit rate} is the fraction of summaries that are cut off by the token limit. The \textbf{continuation overlap} compares what the solver writes after a summary with what it writes after the full context. All four metrics are computed by fixed programs and need no human or model judge. Detailed definitions are provided in
Appendix~\ref{app:summary_metrics}.

\begin{table*}[!htbp]
    \centering
    \caption{
        Summary quality of Base and Trained on the same 512 reasoning contexts per setting, measured by four metrics.
        The contexts are randomly selected.
        Base128 is the untrained summarizer with a 128-token cap. Base and Trained use 512.
        Bold marks the best value among summary conditions.
        Full is the full-context reference.
    }
    \label{tab:summary_mechanical}
    \small
    \setlength{\tabcolsep}{3pt}
    \renewcommand{\arraystretch}{1.05}

    \begin{tabular}{@{}lrrrrrrrrrrrrr@{}}
        \toprule
        Setting
        & \multicolumn{3}{c}{Source-item recall $\uparrow$}
        & \multicolumn{3}{c}{Bad-item rate $\downarrow$}
        & \multicolumn{4}{c}{Continuation overlap $\uparrow$}
        & \multicolumn{3}{c}{Cap-hit rate $\downarrow$} \\
        \cmidrule(lr){2-4}
        \cmidrule(lr){5-7}
        \cmidrule(lr){8-11}
        \cmidrule(l){12-14}
        & Base128 & Base & Trained
        & Base128 & Base & Trained
        & Full & Base128 & Base & Trained
        & Base128 & Base & Trained \\
        \midrule
        Math (4B)
        & .11 & .22 & \textbf{.27}
        & .130 & .295 & \textbf{.068}
        & .543 & .147 & .184 & \textbf{.211}
        & .83 & .69 & \textbf{.19} \\
        Math (R1)
        & .09 & .23 & \textbf{.24}
        & \textbf{.125} & .316 & .155
        & .403 & .215 & .245 & \textbf{.264}
        & .94 & .63 & \textbf{.15} \\
        Zebra
        & .12 & .23 & \textbf{.38}
        & .099 & .207 & \textbf{.067}
        & .540 & .102 & .143 & \textbf{.193}
        & .60 & .62 & \textbf{.05} \\
        Sokoban
        & .35 & .53 & \textbf{.59}
        & .012 & .033 & \textbf{.009}
        & .537 & .152 & .189 & \textbf{.250}
        & .69 & .42 & \textbf{.10} \\
        Futoshiki
        & .16 & .33 & \textbf{.57}
        & .047 & .123 & \textbf{.024}
        & .610 & .152 & .215 & \textbf{.267}
        & .70 & .66 & \textbf{.19} \\
        \bottomrule
    \end{tabular}
\end{table*}

As shown in \cref{tab:summary_mechanical}, Trained improves over Base on all metrics, indicating better summary quality after training. These improvements are expected, since by design a high reward is assigned when the solver's continuation from the summary is close to its continuation from the full reasoning. The summarizer is therefore trained to keep the information that the solver still needs and to avoid adding content that is not in the reasoning. The reward also contains a length penalty, so we check whether writing shorter summaries is enough to get the improvement. Base128 is the untrained summarizer with a 128-token cap. It has the lowest recall and the lowest continuation overlap on every task, so shortness alone does not help. Trained is higher than Base on both, and it also hits the cap far less often. Trained has the best value in every column except the bad-item rate on mathematics with the R1 solver, where Base128 is lower. Appendix~\ref{app:examples} shows Base and Trained summaries written from identical reasoning. 

\paragraph{Reward Ablation}
Notice that although the summary quality is improved in general, as shown in \cref{tab:summary_mechanical}, Base usually has a much higher cap-hit rate, and we want to investigate whether the main improvement comes from following the budget. To isolate the contribution of the Delta-KL term in our reward design, we consider two additional baselines: one where we modify the prompt to let the base summarizer write fewer words (150) while extending the token budget (1{,}024), so that its cap-hit rate is similar to that of Trained, and the other where we directly train the summarizer with only the length penalty for 40 steps. We compare both with Trained, using the Qwen3-1.7B summarizer and the 4B solver, on AIME 2024--2026.

\cref{tab:cap_control} shows the results. With a 150-word requirement in the prompt and a 1{,}024-token cap, the accuracy is similar to that of the original Base, although the cap-hit rate has fallen to 31.8\%, close to that of Trained. The model trained with the length penalty almost never hits the cap (1.7\%), with a median summary length of 139 tokens, but its accuracy deteriorates greatly after training. These results show that following the budget does not automatically give the improvement of Trained, and our reward pushes the model to trade off the necessary information against the budget.

\begin{table}[!htbp]
    \centering
    \caption{Reward ablation on AIME 2024--2026 with the 1.7B summarizer and the Qwen3-4B-Instruct-2507 solver, at a 32k budget with 32 runs per problem. Base (fewer cap hits) asks for 150 words in the prompt and has a cap of 1{,}024 tokens. The other rows ask for 250 words and have a cap of 512 tokens. Median tokens is the median length of a summary. $\Delta$ is the accuracy difference from Trained. The differences are paired by problem, with one standard error.}
    \label{tab:cap_control}
    \small
    \setlength{\tabcolsep}{7pt}
    \renewcommand{\arraystretch}{1.1}
    \begin{tabular}{@{}lrrrr@{}}
        \toprule
        Method & Cap-hit rate & Median tokens & Accuracy & $\Delta$ \\
        \midrule
        Base & 61.6\% & 502 & 55.3 & $-3.9$\pmse{1.2} \\
        Base (fewer cap hits) & 31.8\% & 513 & 54.6 & $-4.7$\pmse{1.2} \\
        Length penalty only & 1.7\% & 139 & 48.1 & $-11.2$\pmse{2.0} \\
        Trained & 24.5\% & 419 & 59.3 & \\
        \bottomrule
    \end{tabular}
\end{table}

\FloatBarrier

\Needspace{6\baselineskip}
\subsection{Scaling to Longer Reasoning Budget}

{We examine whether the gains from local continuation supervision
transfer to longer reasoning horizons.
{To bound state growth at larger budgets, we use a \emph{sink}
variant of append. A sink-$m$ state retains the first summary and up to
$m-1$ most recent subsequent summaries, plus the latest 128-token
verbatim tail. Both Base and Trained use sink8 for mathematics and Zebra, as in the main experiments. Futoshiki and Sokoban retain append mode through 128k. The scaling figures report strict accuracy. A run that uses up its budget without giving an answer is counted as wrong. With this accuracy, the curves of Single, Delethink, Base, and Trained cannot decrease as the budget increases, which makes the figures easier to read. The accuracy with forced answers is reported in Appendix~\ref{app:eval_protocol}.}}

\begin{figure*}[!htbp]
    \centering
    \includegraphics[width=0.73\textwidth]{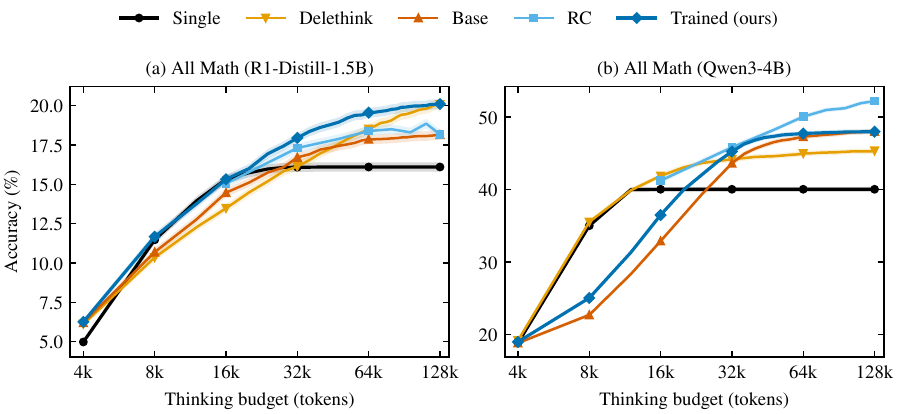}
    \caption{{Math scaling. Accuracy on all 220 mathematics problems (avg@32) for
        (a) a 1.7B summarizer and R1-Distill-1.5B solver, and
        (b) a 4B summarizer and 4B solver. A run that uses up all the tokens of the budget without giving an answer is counted as wrong.}}
    \label{fig:scaling_math}
    \label{fig:scaling_budget}
\end{figure*}

\begin{figure*}[!htbp]
    \centering
    \includegraphics[width=\textwidth]{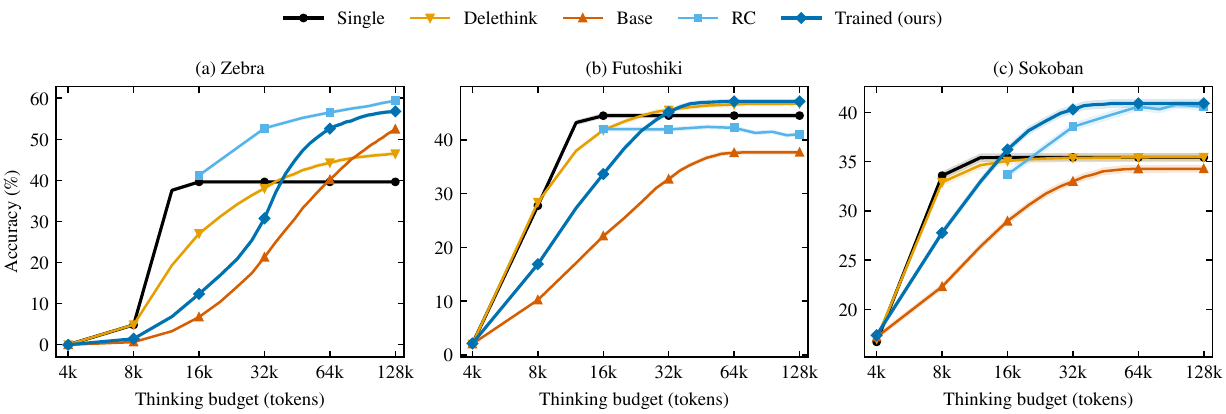}
    \caption{{RG scaling. Accuracy over all
        evaluated difficulty levels of Zebra, Futoshiki, and Sokoban
        with a 4B summarizer and 4B solver (avg@32).
        A run that uses up all the tokens of the budget without giving an answer is counted as wrong.}}
    \label{fig:scaling_rg}
\end{figure*}

{The summarizers are trained on
4k-token reasoning prefixes for only 80 optimization steps, corresponding
to approximately one pass over each
task-specific training set, for both math and Reasoning Gym,
and are evaluated
at larger budgets without further training. As shown in
\cref{fig:scaling_math,fig:scaling_rg}, accuracy improves beyond the point
at which Single plateaus across all evaluated tasks, with gains maintained
at the largest tested budgets. At 128k, Trained exceeds Single by 8.0 percentage points on mathematics with the 4B solver and by 4.0 with the R1 solver. On AIME 2024--2026 alone, the 4B configuration reaches 63.3\% at 128k, exceeding Single by 10.3 percentage points.
On Reasoning Gym the gains are 17.2 points on Zebra, 5.5 on Sokoban and 2.6 on Futoshiki.
These results show that supervision from short, local continuations can
improve reasoning over substantially longer horizons.}

{Trained also improves over Base, although the gap narrows at larger
budgets in some settings. On mathematics with the 4B solver, its advantage decreases from 1.6
percentage points at 32k to zero at 128k. One possible explanation is that
additional reasoning allows the solver to revisit or reconstruct
information omitted by the untrained summarizer, partially compensating
for poorer summaries.
On Futoshiki and Sokoban, however, Base remains below Single even at 128k,
while Trained exceeds Base by 9.4 and 6.6 percentage points, respectively. Finally, RC and Delethink remain two strong baselines in this scaling experiment, we discuss their performance and relationship with other methods in detail in the next subsection.

\begin{samepage}
\subsection{Comparison to strong baselines}\label{sec:baseline_comparison}
 Before we start, we want to highlight that Trained is the best or second-best method in every setting (except HMMT with the R1 solver, where the three methods are within one point), whereas each of the two baselines falls well behind on some task: Delethink on Sokoban and RC on Futoshiki (\cref{tab:main_results,fig:scaling_rg}). Therefore, our method performs consistently well (although not always the best) and is robust in every setting discussed in this paper. We now discuss the pros and cons of Delethink and RC compared with our method.

\par
\end{samepage}

\paragraph{Delethink.}
Delethink only carries the last $M$ tokens of the previous chunk. This is close to copying the tail, so the tokens outside the tail are lost. On Sokoban this lowers accuracy, because the solver has to remember what it established earlier in the trace, such as the coordinate convention and the goal locations. In the trajectories we inspected, the retained suffix often omits them, and the solver can hardly recover them later. A summary can keep this information however early it was established.

\paragraph{RC.}
RC is the strongest baseline in our main benchmark experiments. At 128k, our method outperforms RC on Futoshiki (+6.1) and on mathematics with the R1 solver (+1.9), and RC outperforms ours on mathematics with the 4B solver (+4.2) and on Zebra (+2.6). On these two tasks the curve of RC has not plateaued at 128k, so its advantage may grow at larger budgets.

First, RC needs each attempt to have a fair probability of solving the problem on its own. RC runs complete attempts of up to 16k tokens one after another and carries a cache between them, so the progress of the attempts adds up. This works well when a single attempt already solves the problem with a fair probability, as on mathematics. It works poorly when the context of a single attempt grows so long that the solver collapses and almost never solves some part of the problem. The cache then has little to carry, and the next attempt fails at the same place for the same reason. Our method cuts the trace inside an attempt and keeps every context short, so it raises the accuracy of each attempt.

The multiplication task of \cref{sec:motivation} shows this (\cref{fig:rc_mul}). We give RC unfair advantages. It is given a 128k budget, while the other methods use 49k, and an oracle cache that lists every problem the solver has answered together with its answer. Consider Problems 57 to 63. In the first attempt the solver reaches them in 53\% of the runs, and only 7\% of its answers are correct, because these problems come after about 11k tokens of context. In each later attempt it answers them in about 20\% of the runs, and fewer than half of these answers are correct. For each attempt, the accuracy on these problems is 0.04 after the first attempt, 0.12 after the second, and 0.28 after the eighth, about 3 points per attempt. Trained reaches 0.80 on these problems in a single run with a 49k budget. Appendix~\ref{app:synthetic} gives the setup and the results after each attempt (\cref{tab:rc_mul_attempts}).

\begin{figure}[!htbp]
    \centering
    \includegraphics[width=0.6\textwidth]{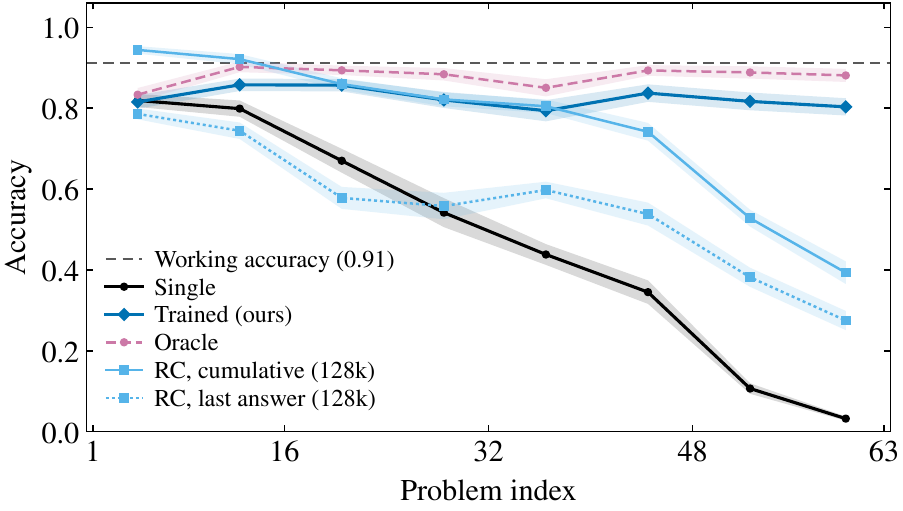}
    \caption{Accuracy of RC on the 64 multiplication problems, averaged in groups of eight. RC uses a 128k budget and an oracle cache. The other curves are those of \cref{fig:e0}, with a 49k budget. Last answer scores the last answer that the solver gave to each problem. Cumulative counts a problem as solved if any attempt answers it correctly.}
    \label{fig:rc_mul}
\end{figure}

Second, RC is the only method in our comparison that continues to reason after the solver has given an answer. All the other methods stop once the solver outputs \texttt{[EOS]} and gives an answer, while RC makes new attempts until the budget is used up. So RC does not run in the same mode as the other methods. The extra attempts help in most settings. On mathematics with the 4B solver at 128k, RC reaches 42.9\% if it stops at the first attempt that ends on its own, and 52.2\% with all eight attempts. On some tasks, however, they hurt. On Futoshiki an erroneous ``no solution'' conclusion passes through the cache and is repeated in later attempts, and a correct earlier answer can be replaced by a later wrong one. Stopping at the first complete answer would give RC 49.9\% at 128k, against 47.2\% for Trained.

Finally, we think the two approaches are complementary. \citet{wu2026reasoningcache} discuss the credit assignment problem of training the cache with the outcome reward. Delta-KL is a training signal for the state that does not depend on the outcome reward, so it could also supervise the cache of RC. RC's 16k attempts lie outside the short contexts in which our reference continuations are reliable, so this would require adapting the training setup.

\subsection{Efficiency and Computation}

\paragraph{Inference efficiency.}
\cref{fig:inference_efficiency} reports the average inference cost per problem with the Qwen3-4B-Instruct solver on one H100 GPU. The cost of a run is counted up to the point where it stops. With sink8, which retains at most eight summaries, TruSuCo keeps the peak KV cache per sequence at about 0.7 GB at every budget, comparable to Delethink, while the memory of Single grows with the budget. At small budgets TruSuCo costs more than Single, because Single often stops early and TruSuCo continues to reason. Its cost levels off as the budget grows. At 128k, TruSuCo takes 1.9 minutes and 0.49 PFLOPs per problem, against 2.2 minutes and 0.70 PFLOPs for Single and 3.7 minutes and 0.70 PFLOPs for RC. Delethink is the cheapest of the four methods. The summarizer is a separate model and adds about 8 GB of weights. Appendix~\ref{app:inference_compute} describes how the costs are measured.

\begin{figure*}[!htbp]
    \centering
    \includegraphics[width=0.98\textwidth]{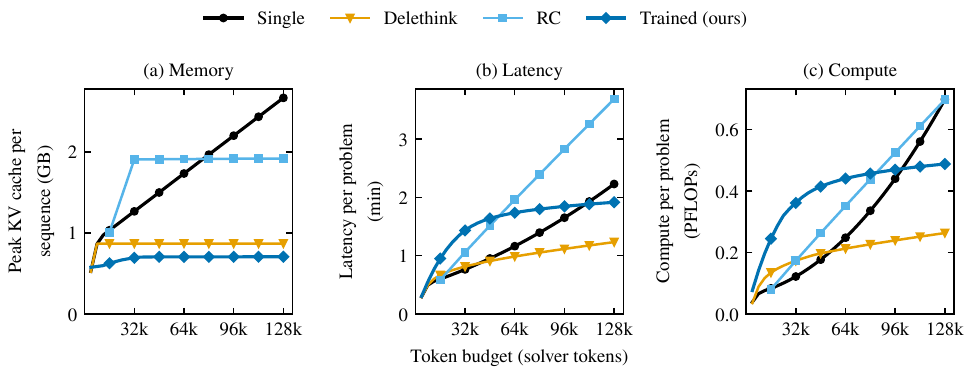}
    \caption{Average inference cost per problem with the Qwen3-4B-Instruct-2507 solver on
        one H100 GPU as a function of the solver-token budget, at batch size 1.
        The cost of a run is counted up to the point where it stops.
        (a) Peak KV cache per sequence, taking the maximum over
        solver and summarizer calls. Model weights are excluded, and
        TruSuCo's separate summarizer adds approximately 8 GB of weights.
        (b) End-to-end latency per problem, including summarizer calls.
        (c) Compute per problem in PFLOPs ($10^{15}$ floating-point operations),
        including summarizer calls. TruSuCo uses sink8.}
    \label{fig:inference_efficiency}
\end{figure*}

\paragraph{Training efficiency.}
For 80 optimization steps, the 4B summarizer paired with a frozen 4B solver takes approximately 5--6 hours on three H100 GPUs, with two GPUs for summarizer training and one for solver scoring. The 1.7B summarizer paired with R1-Distill-1.5B takes approximately 4 hours on two H100 GPUs, one for training and one for scoring. Detailed training costs are provided in Appendix~\ref{app:training_costs}.

\section{Related Work}

\paragraph{Context management with a textual state.}
Several recent methods run the reasoning model in segments and carry a textual state from one segment to the next. In some methods the state is a summary that the model writes itself~\citep{yan2025inftythink,yan2026inftythinkplus,wu2026reasoningcache,kontonis2026memento,yang2026accordion,wang2026resum}. Other methods keep part of the trace. Delethink keeps the last tokens of the previous segment~\citep{aghajohari2025markovian}, and PENCIL and Free()LM let the model delete the reasoning that it no longer needs~\citep{yang2025pencil,zheng2026freelm}. Agents do something similar with their interaction history. MEM1 consolidates it into a compact state at every turn~\citep{zhou2025mem1}, and MemGPT pages it in and out of an external memory~\citep{packer2023memgpt}. Other agents replace the earlier history with a summary, and the compaction policy is trained by the task reward~\citep{li2026compactionrl}, adapted through counterfactual continuations of the task~\citep{min2026pair}, or only prompted~\citep{li2026selfcompacting}. Compression into text has also been studied for prompts and retrieved documents. LLMLingua drops the tokens that carry little information under a small language model~\citep{jiang2023llmlingua}. RECOMP~\citep{xu2024recomp} and TaC~\citep{ma2026tac} train the compressor so that a downstream model answers correctly from the compressed text. RECOMP uses supervised learning, and TaC uses RL with a task reward. When these systems are trained, the signal is usually the task outcome or a target produced by a stronger model. In our method a separate summarizer writes the state, and it is trained on the continuation behavior of the frozen solver.

\paragraph{Dedicated learning signals for context compression.}
A second line of work gives the compression a dedicated training objective. Most of these methods compress the context into a few continuous vectors, such as gist tokens, summary embeddings, or compressed key-value entries~\citep{mu2023gist,chevalier2023autocompressors,ge2023icae,kim2023ccm,chari2025kvdistill}. The training objective is a language-modeling or KL-type loss on the outputs of the model that reads the compressed representation. The representation is continuous, so the compressor is trained by backpropagating this loss through that model. When the compressed context is text, this gradient is not available. RL has also been used to compress instruction prompts for a frozen model. PCRL~\citep{jung2023pcrl} and TACO-RL~\citep{shandilya2024tacorl} train a policy to delete tokens from the prompt. The model generates one output from the original prompt and one from the compressed prompt, and the reward is the similarity between the two outputs. Among these works, the following three appear closest to our method, but each differs from it in a fundamental way.
\begin{itemize}
    \item \textbf{REMORY}~\citep{xia2026remory} is a concurrent work. It appends learned soft tokens to a fixed textual summary and trains them with a KL divergence between a frozen model's next-token distributions under the full history and under the summary with the tokens. Our criterion coincides with this objective, but our state is text. The gradient cannot pass through text, so we turn the criterion into a reward and train the summarizer with RL.
    \item \textbf{CoMem}~\citep{zhang2026comem} trains a separate memory model with GRPO to summarize the history of a frozen agent. Its reward is the similarity between the action that the agent takes from the summary and the action that it takes from the full history. Their goal is to reduce latency. In contrast, our reward is a reweighted KL estimate computed from token-level log-probabilities, and our goal is to self-improve the long-horizon reasoning capability.
    \item \textbf{MaxIM}~\citep{jeong2026maxim} is a concurrent work. It also trains a text summarizer with RL, but it keeps outcome supervision. A critic scores how much each summary helps predict targets produced by another LLM, and this score shapes the task reward. Our reward needs no outcome and is computed by the frozen solver alone.
\end{itemize}

\paragraph{Long-horizon reasoning.}
Using more tokens at test time improves reasoning~\citep{snell2024scaling,muennighoff2025s1}. But models degrade as their context grows, even when the relevant information is still present~\citep{liu2024lost,hsieh2024ruler,modarressi2025nolima,du2025context}, so a longer trace does not by itself give a longer effective horizon. h1~\citep{motwani2025h1} constructs training problems with a Markovian structure, in which short problems are chained and each one depends on the answer of the previous one. It trains the model on these problems with curriculum RL and improves its long-horizon reasoning. \citet{yuan2026markovstates} show that training the model with RL on compact Markov states, in place of the full action history, gives an exponential improvement in sample complexity and raises pass@$k$, so the model finds solutions that it could not reach before. Our summarizer can be viewed as an estimator of the Markov state of the reasoning process. We keep the solver frozen in this work, and we leave RL training of the solver on top of the summarizer and TruSuCo to future work.

\section{Conclusion}
In this paper we studied how to improve LLM reasoning over long horizons by periodically replacing its reasoning history with a compact text summary. We trained the summarizer with Delta-KL, a reward computed from the frozen solver's own continuation distributions, so that the solver reasons from the summary as it would from the full history. Training used a few thousand questions and no answers, verifiers, or stronger teacher. On mathematical benchmarks and Reasoning Gym puzzles under a 32k-token budget, the trained summarizer improved accuracy over vanilla reasoning and an untrained summarizer. The improvement increased with the reasoning budget and scaled to a 128k-token budget, even though the summarizer was only trained on 4k-token reasoning traces. 

\clearpage
\bibliographystyle{plainnat}
\bibliography{ref}

\clearpage
\appendix
\section{Synthetic Task Details}\label{app:synthetic}

\paragraph{Question.}
Each instance of the synthetic task of \cref{sec:motivation} is the following question, with 64 problems that each multiply a three-digit number by a two-digit number:
\begin{promptframe}
You will solve 64 word problems in order. Every answer is a whole number.\\
For each problem: solve it, then write its answer on its own line as \textquotedbl Answer k = <value>\textquotedbl.\\
The final answer is the answer to Problem 64; give it in \textbackslash boxed\{\}.\\[3pt]
Problem 1:\\
What is 730 $\times$ 74?\\[3pt]
\ldots\\[3pt]
Problem 64:\\
What is 446 $\times$ 78?
\end{promptframe}

\paragraph{Solver prompt.}
The prompt of the solver is this question followed by the solver's standard instruction, wrapped in its chat template:
\begin{promptframe}
[the question above]\\
Please reason step by step, and put your final answer within \textbackslash boxed\{\}.
\end{promptframe}

\Needspace{30\baselineskip}
\paragraph{Summarizer prompt.}
Base and Trained receive the following prompt. The braced fields are replaced by the corresponding text.
\begin{Verbatim}[fontsize=\small,breaklines=true,breakanywhere=false,breakindent=0pt,breaksymbolleft={},frame=single,framesep=6pt]
Below is a maths problem and the beginning of my own reasoning about it. The reasoning was cut off partway, so it may stop mid-thought.

Rewrite my reasoning as a much shorter version of itself, in the same voice and style as the original: first person, thinking out loud ("So,", "Let me", "Hmm,", "Wait,"), plain sentences and short lines, and the same notation (LaTeX with \( \)). Start with a few words on how far I have got. Then continue with the last part of my reasoning, copied exactly, up to the point where it stops. Copy every label, name and number exactly as I wrote it, and never add a label I did not write. Drop false starts, repeated checks and explanations.

Do not add anything that is not in my reasoning, do not take the work further, and do not fix anything that looks wrong.

Do not describe the reasoning from the outside: no "the attempt", "the work stopped", "I have established", no notes, headings, bullet summaries or bold text. It must read as if I had simply thought more concisely from the start.

End your rewrite with my last line, copied exactly as shown under MY LAST LINE, even though it is cut off mid-way. Use at most 45 words.

### PROBLEM
{problem}

### MY REASONING SO FAR
{reasoning}

### MY LAST LINE
{last_line}
\end{Verbatim}
From the second cut on, the prompt also contains the previous summary in a section headed \texttt{\#\#\# EARLIER}, with one more sentence of instruction.

\Needspace{16\baselineskip}
\paragraph{Oracle state.}
The Oracle state has a fixed format. It first says which problems are done. It then lists the solver's most recent answers, as many as fit within the 128 tokens, and ends with the problem that the solver is working on. For the run shown at the end of this appendix, the first state reads:
\begin{promptframe}
Okay, so let me recap where I am.\\[3pt]
Done with Problems 1 to 12.\\
Problem 6: 47,160.\\
Problem 7: 60,606.\\
Problem 8: 16,160.\\
Problem 9: 62,464.\\
Problem 10: 16,685.\\
Problem 11: 51,410.\\
Problem 12: 22,287.\\[3pt]
Working on Problem 13: What is 667 $\times$ 34?
\end{promptframe}

\paragraph{Training and inference.}
We train the summarizer for this task with the procedure of Section~\ref{sec:methodology} and different hyperparameters. We use a window of $K=2{,}048$ tokens, a tail of $k=128$ tokens, and a cap of 128 tokens on each summary. At inference, we use the roll mode of Section~\ref{sec:methodology}. At each cut, the summarizer reads the question and the first 1{,}920 tokens of the window (as well as previous summary, if there is), and the last 128 tokens are retained verbatim. 

\paragraph{Scoring.}
We only score the text generated by the solver. The parser accepts several ways of stating the answer to Problem $k$: ``Answer $k$ = $v$'', ``Problem $k$'s answer is $v$'', ``the answer to Problem $k$ is $v$'', ``Problem $k$ ($v$)'', and the last integer of a section headed ``Problem $k$:''. For the last problem it also accepts the number after ``Final Answer'' or inside \texttt{\textbackslash boxed\{\}}.  We keep the first answer the solver gives for each problem and ignore later revisions.  We average the scores over the 16 runs of a question and then over the 24 questions. The working accuracy in \cref{fig:e0} is the accuracy on the problems that the solver answers before the first cut, pooled over the runs of all six methods.

\paragraph{Additional statistics.}
We also count how often a summary says which problems are done. We look for five patterns, such as ``Done with Problems 1 through 8'', ``I've gone through Problems 1 through 9'' or ``Problems 49--52 done'', and take the last problem named as the position the summary states. We compare it with the last problem the solver had completed 128 tokens before the cut, which is where the retained tail begins. A problem is completed at that point if the answer written for it is the one the solver finally gives. The stated position is correct if it is within one problem of this reference, because the boundary can fall inside a problem. \cref{tab:mul_progress} shows the result for all summaries of the runs in \cref{fig:e0}. The trained summarizer states the position in 90.0\% of its summaries and the untrained one in 51.8\%. The stated position is correct in 84.0\% of the trained summaries and in 46.4\% of the untrained ones. The runs also end differently. With the untrained summarizer, 17.4\% of the runs end before the solver reaches Problem 64, against 11.5\% with the trained one. In about half of these runs with the untrained summarizer the solver gives the answer to the problem it is working on as its final answer, as in the example below. This happens in 8.6\% of all runs with the untrained summarizer and in 1.0\% with the trained one.

\begin{table}[!htbp]
    \centering
    \caption{Statements of progress in the summaries of the synthetic task, for the runs in \cref{fig:e0}. Percentages are shares of all summaries of a method.}
    \label{tab:mul_progress}
    \small
    \setlength{\tabcolsep}{8pt}
    \renewcommand{\arraystretch}{1.15}
    \begin{tabular}{@{}lrr@{}}
        \toprule
        & Base & Trained \\
        \midrule
        Number of summaries & 1{,}679 & 2{,}209 \\
        States which problems are done & 51.8\% & 90.0\% \\
        Stated position is correct & 46.4\% & 84.0\% \\
        \bottomrule
    \end{tabular}%
    
\end{table}

\paragraph{RC with an oracle cache.}
\cref{fig:rc_mul} uses RC with eight attempts of at most 16{,}384 tokens, as in the main experiments, on the same 24 questions with 16 runs each. The first attempt uses the solver prompt above. The later attempts use the reasoning prompt of Reasoning Cache, with one added sentence saying that every result the question asks for is part of the answer. The cache is not written by a model. It is written by a rule, which lists the latest answer that the solver wrote for each problem in the previous attempts. \emph{Last answer} takes, for each problem, the answer of the most recent attempt that answered it, and a problem that no attempt answered counts as wrong. \emph{Cumulative} counts a problem as correct once any attempt has answered it correctly. \cref{tab:rc_mul_attempts} gives the results on Problems 57 to 63 for each attempt. In the first attempt the answers to these problems come after a median of 11.0k tokens.

\begin{table}[!htbp]
    \centering
    \caption{RC with an oracle cache on Problems 57 to 63 of the synthetic task. The second column is the share of these problems that the solver answers in the attempt. The last two columns give the accuracy after the attempt over all runs and problems, where a problem without an answer counts as wrong. Trained reaches 0.80 on these problems in a single run.}
    \label{tab:rc_mul_attempts}
    \small
    \setlength{\tabcolsep}{10pt}
    \renewcommand{\arraystretch}{1.12}
    \begin{tabular}{@{}cccc@{}}
        \toprule
        Attempt & Share answered in the attempt & Last answer & Cumulative \\
        \midrule
        1 & 0.53 & 0.04 & 0.04 \\
        2 & 0.21 & 0.12 & 0.12 \\
        3 & 0.23 & 0.17 & 0.19 \\
        4 & 0.20 & 0.21 & 0.25 \\
        5 & 0.15 & 0.22 & 0.28 \\
        6 & 0.22 & 0.24 & 0.32 \\
        7 & 0.18 & 0.26 & 0.36 \\
        8 & 0.20 & 0.28 & 0.40 \\
        \bottomrule
    \end{tabular}
\end{table}

\paragraph{Example.}
The boxes below show one instance. The two runs share the first chunk, which is cut in the middle of Problem 13. The base summary only keeps the current product. The solver completes it, treats it as the whole task, writes a final answer and stops after 3{,}050 tokens, so Problems 14 to 64 are never attempted. The trained summary also states which problems are done. The solver completes Problem 13 and moves on to the next problems, and it gets 63 of the 64 problems right after 6 summaries and 13{,}568 tokens. The Oracle statement above is written at the same cut and names the same problems as done. With Oracle the solver gets 60 of the 64 problems right after 5{,}888 tokens. The last two boxes show RC with the oracle cache on the same question and sample. The cache given to its eighth attempt has an answer for every problem, and 34 of the 64 answers are correct. Even with these answers in the cache, the solver starts again from Problem 1 and solves every problem again. Its context then contains the cache and its new reasoning, and the solver collapses again. From about token 8{,}800 it goes through Problems 41 to 64 with the numbers of other problems. It replaces 5 correct answers with wrong ones and corrects 5 wrong ones, and the run ends with 34 of the 64 problems right after eight attempts. In the last box, the number in brackets is the number of tokens generated in the attempt. The gray lines that start with ``Our note'' are written by us and are not part of the solver's output. Texts are shown as generated, and \exomit{} marks text that we omit.

\medskip
\begin{exbox}[\linewidth]{End of the first chunk, identical in both runs}
\exomit{}\par
Problem 13: 667 \exsym{\times} 34. Breaking this down: 667 \exsym{\times} 30 is 20,010 and 667 \exsym{\times} 4 is 2,668. Adding them together: 20,010 + 2,668 = 22,678. Alternatively, 667 \exsym{\times} 34 is (600 + 67) \exsym{\times} 34 = 600\exsym{\times}34 + 67\exsym{\times}34. 600\exsym{\times}34 is 20,400 and 67\exsym{\times}34 is 2\par
\end{exbox}\par\medskip
\noindent
\begin{exbox}[0.492\textwidth]{\textcolor{black!65}{Base summary 1}}
Okay, so let me recap where I am.\par
\exgap
Problem 13: 667 \exsym{\times} 34.\par
\end{exbox}\hfill
\begin{exbox}[0.492\textwidth]{\textcolor{Teal}{Trained summary 1}}
Okay, so let me recap where I am.\par
\exgap
Okay, so I've gone through Problems 1 through 12, breaking each multiplication down using distributive properties and step-by-step addition. All answers are verified via two methods where possible.\par
\exgap
Problem 13: 667 \exsym{\times} 34.\par
\end{exbox}
\par\medskip
\noindent
\begin{exbox}[0.492\textwidth]{\textcolor{black!65}{Base: the solver's continuation (end of the run)}}
,278. So, 20,400 + 2,278 = 22,678.\par
\exomit{}\par
Therefore, I feel confident that the answer to Problem 13 is 22,678.\par
\exgap
**Final Answer**\par
\textbackslash{}boxed\{22678\}\par
</think>\par
\exomit{}\par
\end{exbox}\hfill
\begin{exbox}[0.492\textwidth]{\textcolor{Teal}{Trained: the solver's continuation (start)}}
,278. Adding those gives 22,678. Seems consistent. So, Problem 13: 22,678.\par
\exgap
Problem 14: 355 \exsym{\times} 41.\par
\exgap
Hmm, 355 \exsym{\times} 40 is 14,200 and 355 \exsym{\times}1 is 355. So total is 14,200 + 355 = 14,555. Alternatively, 355 \exsym{\times}41 = (300 +55) \exsym{\times}41 = 300\exsym{\times}41 +55\exsym{\times}41. 300\exsym{\times}41 is 12,300 and 55\exsym{\times}41 is 2,255. Adding those gives 14,555. Correct.\par
\exgap
Problem 15: 318 \exsym{\times}31.\par
\exomit{}\par
\end{exbox}
\par\medskip

\noindent
\begin{exbox}[0.492\textwidth]{RC: the cache given to the eighth attempt}
These are the answers I have obtained so far in my previous attempts:\par
Answer 1 = 54020\par
Answer 2 = 24420\par
Answer 3 = 18360\par
\exomit{}\par
Answer 62 = 56576\par
Answer 63 = 7514\par
Answer 64 = 34788\par
\exgap
Problems without an answer yet: none.\par
\end{exbox}\hfill
\begin{exbox}[0.492\textwidth]{RC: the eighth attempt}
[token 0] Alright, so I've got this problem set with 64 questions, each requiring a whole number answer.\par
\exomit{}\par
[token 8,761] Problem 41: 220 \exsym{\times} 70. Let me compute this again. 220 \exsym{\times} 70 is 15,400. Correct.\par
{\normalfont\itshape\color{black!55}[Our note: Problem 41 is 838 $\times$ 64, and the cache has the correct answer 53,632.]}\par
\exgap
[token 9,098] Problem 44: 869 \exsym{\times} 60. Let me compute this again. 800 \exsym{\times} 60 is 48,000; 60 \exsym{\times} 9 is 540. Adding those gives 48,540. Correct.\par
{\normalfont\itshape\color{black!55}[Our note: Problem 44 is 557 $\times$ 51, and the cache has the correct answer 28,407.]}\par
\end{exbox}
\par\medskip

\section{Reasoning Gym Tasks}\label{app:rg_tasks}

{Reasoning Gym~\citep{stojanovski2025reasoninggym} is a library of reasoning tasks whose problems are produced by programs. Each task has a generator and a verifier. The generator takes a configuration, which is a set of parameters that control the size and the difficulty of the puzzles, and a random seed.  The verifier checks whether an answer solves a given puzzle. We use the three tasks described below, and Appendix~\ref{app:examples} shows an example of each task.}

\paragraph{{Zebra.}}
{Several people live in a row of houses, one person in each house. Each person has several characteristics, such as a name, a favorite color and a pet, and no two people share a value. The puzzle lists clues that relate these characteristics, for example that one person lives directly left of the owner of a certain pet. The question asks for the name of the person who lives in a given house. The configuration sets the number of people and the number of characteristics.}

\paragraph{{Sokoban.}}
{The puzzle is a grid with walls, boxes, goals and a player, written with one character per cell. The player moves one cell at a time, up, down, left or right, and pushes a box by walking into it. The puzzle is solved when every box is on a goal. The answer is the sequence of moves, written as a string of the letters U, D, L and R. The configuration sets the range of the width and the height of the grid and the range of the number of boxes. It also sets \texttt{max\_depth}, the search depth within which the solver of the generator has to find a solution for a puzzle to be kept.}

\paragraph{{Futoshiki.}}
{The puzzle is an $n\times n$ grid in which some cells are filled with numbers. The grid has to be completed so that every row and every column contains each of the numbers $1$ to $n$ exactly once. Some pairs of neighboring cells carry an inequality sign, and the completed grid has to satisfy all of them. The answer is the completed grid. The configuration sets the grid size and a difficulty parameter between 0 and 3, and a larger value places more inequality signs in the grid.}

\paragraph{{Difficulty levels and data.}}
{The labels L0 to L3 are our names for configurations of increasing difficulty within a task. \cref{tab:rg_test_configs} lists the configurations and seeds of the test sets. Each task is tested at two levels with 96 puzzles per level, which gives the 192 test puzzles per task. Training uses the same generators with 1{,}200 puzzles per level. For Zebra the training levels are L1 (6 people, 5 characteristics) and L2 (7 people, 6 characteristics), so the test level L3 has one more characteristic than any puzzle seen in training. For Sokoban and Futoshiki the training levels are L0 and L1, with the configurations in the table. Appendix~\ref{app:eval_protocol} describes how the answers are scored.}

\begin{table*}[!htbp]
    \centering
    \caption{{Reasoning Gym test-set generation settings.}}
    \label{tab:rg_test_configs}
    \small
    \setlength{\tabcolsep}{5pt}
    \renewcommand{\arraystretch}{1.15}
    {%
    \begin{tabular*}{\textwidth}{@{\extracolsep{\fill}}llp{0.48\textwidth}r@{}}
        \toprule
        Task & Level & Configuration & Seed \\
        \midrule
        Zebra & L2 & 7 people, 6 characteristics & 777002000 \\
        Zebra & L3 & 7 people, 7 characteristics & 777003000 \\
        Sokoban & L0 & Width and height 6--7, 2--3 boxes, \texttt{max\_depth}=40 & 777000000 \\
        Sokoban & L1 & Width and height 7--8, 3--4 boxes, \texttt{max\_depth}=60 & 777001000 \\
        Futoshiki & L0 & $5\times5$ grid, difficulty {parameter} 1--2 & 777000000 \\
        Futoshiki & L1 & $6\times6$ grid, difficulty {parameter} 2--3 & 777001000 \\
        \bottomrule
    \end{tabular*}%
    }
\end{table*}

\section{Training and Inference Implementation Details}\label{app:implementation}

\subsection{Inference}\label{app:inference_details}

{Table~\mbox{\ref{tab:inference_hparams}} lists the inference settings for the main experiments.}

\begin{table}[!htbp]
    \centering
    \caption{{Inference and generation settings for TruSuCo in the main experiments.}}
    \label{tab:inference_hparams}
    \small
    \setlength{\tabcolsep}{5pt}
    \renewcommand{\arraystretch}{1.12}
    {%
    \begin{tabular}{@{}p{0.43\textwidth}p{0.52\textwidth}@{}}
        \toprule
        Setting & Value \\
        \midrule
        Reasoning window length $K$ & 4,096 \\
        Retained tail length $k$ & 128 \\
        Summary cap, including the fixed opening & 512 \\
       
        Retention for mathematics and Zebra & Sink8: the first summary and up to seven most recent subsequent summaries \\
        Retention for Sokoban and Futoshiki & Append: all summaries \\
        \midrule
        Qwen3-4B-Instruct-2507 solver sampling & Temperature 0.7, top-$p$ 0.8, top-$k$ 20 \\
        R1-Distill-Qwen-1.5B solver sampling & Temperature 0.6, top-$p$ 0.95 \\
        Summarizer sampling at inference & Temperature 0.7, top-$p$ 0.8, top-$k$ 20 \\
        \bottomrule
    \end{tabular}%
    }
\end{table}

\paragraph{Reasoning windows and summary retention.}
In the main experiments, the solver first generates up to 4,096 tokens. If it has not finished and still has budget left, we cut the trace. The summarizer reads the question and the first 3,968 solver tokens of the local window. The last 128 tokens are retained verbatim. After each cut, the solver generates up to 3,968 new tokens, so the next window consists of the retained tail followed by the new reasoning. The previous tail is therefore included in the next material to summarize. Thus each reasoning token is summarized once when it leaves the window, and only the latest tail remains in the solver's state. Neither append nor sink includes previous summaries in the summarizer's input.

Append retains all summaries. Sink8 retains the first summary and the latest seven subsequent summaries, in their original order, or all summaries if fewer than eight exist. The nominal content budget for sink8 is therefore $8\times512+128=4{,}224$ tokens. Append has no fixed total-state limit. 

\paragraph{Summary formatting and length control.}

Every summary begins with ``Okay, so let me recap where I am.'' We allow up to 500 generated tokens so that this opening and the generated text fit within the 512-token budget. We put a blank line (\texttt{\textbackslash n\textbackslash n}) between successive summaries and another before the retained tail. The solver continues from the end of that tail. In the summarizer's prompt, we also ask it to write at most 250 words.

\FloatBarrier

\subsection{Training}\label{app:training_details}

We only train the summarizer, keeping the solver and the stored probes fixed. The window and tail lengths and the summary limit are the same as at inference (Table~\ref{tab:inference_hparams}). Table~\ref{tab:training_hparams} gives the optimization and reward settings. All training runs use the same settings.

\begin{table}[!htbp]
    \centering
    \caption{{Training and reward computation settings for the main experiments.}}
    \label{tab:training_hparams}
    \small
    \setlength{\tabcolsep}{5pt}
    \renewcommand{\arraystretch}{1.12}
    {%
    \begin{tabular}{@{}p{0.49\textwidth}p{0.46\textwidth}@{}}
        \toprule
        Setting & Value \\
        \midrule
        Update algorithm & GRPO \\
        Training framework & verl \\
        Stored continuation probes per prefix $J$ & 8 \\
        Maximum probe length & 1,024 solver tokens \\
        Candidate summaries per question $G$ & 8 \\
        Questions per batch & 32 \\
        Optimization steps & 80 \\
        Gradient updates per step & 1, on all 256 summaries of the batch \\
        Optimizer & AdamW, weight decay 0.01 \\
        Learning rate & $10^{-6}$, constant \\
        Gradient clipping & 1.0 \\
        Trained parameters & All parameters of the summarizer \\
        GRPO clip range & 0.2 \\
        KL regularization toward the initial summarizer & Coefficient 0.001 \\
        Summarizer sampling in training & Temperature 1.0, top-$p$ 1.0 \\
       
        Reference length for token weighting & 640 solver tokens \\
        Cap-hit penalty coefficient $\lambda$ & 1.0 \\
        \bottomrule
    \end{tabular}%
    }
\end{table}

\paragraph{Reference for token weighting.}

We form $s^{\mathrm{tail}}$ by copying the last 640 tokens of the reasoning. This matches the nominal budget of one summary plus the retained tail. We only sample probes from the full reasoning, then evaluate the same tokens' log-probabilities under the full reasoning and the copied suffix. 

\paragraph{Continuation probes.}

We score each probe from its first token, using at most the first 1,024 generated tokens (Table~\ref{tab:training_hparams}). If it ends sooner, we use the tokens it has generated. Probes whose token weights sum to zero are left out of the average. In our training data, every question has at least one probe with positive total weight, so \eqref{eq:rdkl} is always defined.

\section{Prompts for the Main Experiments}\label{app:main_prompts}

{The following prompts are used for mathematics and Reasoning Gym. Each model uses its own tokenizer and native chat template, with a single user message and no system message. Braced fields are replaced by the corresponding input text.}

\paragraph{Solver prompt.}
{The solver receives the original question followed by the same fixed instruction. Reasoning Gym questions retain the generator's text and answer-format instructions.}
\begin{Verbatim}[fontsize=\small,breaklines=true,breakanywhere=false,breakindent=0pt,breaksymbolleft={},frame=single,framesep=6pt]
{question}
Please reason step by step, and put your final answer within \boxed{}.
\end{Verbatim}

\Needspace{32\baselineskip}
\paragraph{Summarizer prompt.}
{ The field \texttt{\{problem\}} is the original question, \texttt{\{reasoning\}} is the part of the local window preceding the retained tail, and \texttt{\{last\_line\}} is the last line of that text, limited to 300 characters. Earlier summaries are not included.}
\begin{Verbatim}[fontsize=\small,breaklines=true,breakanywhere=false,breakindent=0pt,breaksymbolleft={},frame=single,framesep=6pt]
Below is a maths problem and the beginning of my own reasoning about it. The reasoning was cut off partway, so it may stop mid-thought.

Rewrite my reasoning as a much shorter version of itself, in the same voice and style as the original: first person, thinking out loud ("So,", "Let me", "Hmm,", "Wait,"), plain sentences and short lines, and the same notation (LaTeX with \( \)). Keep the results and quantities I will still need, copied exactly, each with a short account of how I got it; what I have finished with can be shortened to a line. Drop false starts, repeated checks and explanations.

Do not add anything that is not in my reasoning, do not take the work further, and do not fix anything that looks wrong.

Do not describe the reasoning from the outside: no "the attempt", "the work stopped", "I have established", no notes, headings, bullet summaries or bold text. It must read as if I had simply thought more concisely from the start.

End your rewrite with my last line, copied exactly as shown under MY LAST LINE, even though it is cut off mid-way. Use at most 250 words.

### PROBLEM
{problem}

### MY REASONING SO FAR
{reasoning}

### MY LAST LINE
{last_line}
\end{Verbatim}

\paragraph{Chat formatting and continuation.}
{Qwen3-4B-Instruct-2507 opens a plain assistant turn. The R1-Distill-Qwen-1.5B solver's template opens a \texttt{<think>} block. For the Qwen3-1.7B summarizer, thinking is disabled and the template supplies \texttt{<think>} and \texttt{</think>} with no reasoning between them.}

{Before the solver sees a summary, the fixed opening below is prepended to the generated text. The summaries and retained tail are inserted directly into the solver's open assistant turn, using the separators and retention rules in Appendix~\ref{app:implementation}.}
\begin{Verbatim}[fontsize=\small,breaklines=true,breakanywhere=false,breakindent=0pt,breaksymbolleft={},frame=single,framesep=6pt]
Okay, so let me recap where I am.

{summary}
\end{Verbatim}

\section{Evaluation Protocol Details}\label{app:eval_protocol}

\paragraph{Budget and termination.}
The budget only counts the tokens that the solver generates, summed over all chunks. With DeepSeek-R1-Distill-Qwen-1.5B the 128k budget stops at 126{,}976 tokens, the limit of the model's position window. A run terminates naturally when the solver emits \texttt{</think>} or \texttt{[EOS]} within the budget. After \texttt{</think>} it may write up to 1{,}024 answer tokens. Qwen3-4B-Instruct-2507 has no reasoning block and does not write \texttt{</think>}, so its runs only terminate naturally with \texttt{[EOS]}. If the budget runs out first, we force an answer. We append the line ``\texttt{I have run out of thinking budget, so I will stop here and give my final answer.}'' and \texttt{</think>}, and the solver then writes its answer within 1{,}024 tokens. We only score the text after \texttt{</think>}, so a forced run is judged on its forced answer.

For the 128k runs we also need the accuracy at smaller budgets. Every 8{,}192 tokens we pause each unfinished run and apply the forcing procedure to a copy of its context. We store the answer and let the run continue. One run then gives the accuracy at 8k, 16k, \ldots, 128k.

\paragraph{Baseline configurations.}
\emph{RC.} Attempts have 16{,}384 tokens and their number is the budget divided by 16k (two at 32k, eight at 128k; for R1 at 124k, eight attempts of 15{,}872). The cache is written with the official Reasoning Cache prompt, only changed from ``a maths problem'' to ``a problem'', and capped at 2{,}048 tokens.  The 4B solver writes its own cache, as in Reasoning Cache. R1-Distill-1.5B produces a thinking trace from which no usable cache can be written, so the untrained Qwen3-1.7B writes it with the same prompt. RC runs all its attempts and the last answer is scored. For RC, an attempt is only forced to answer when it hits its 16k cap. On mathematics with the 4B solver at 128k, the first attempt ends on its own in 6{,}393 of the 7{,}040 runs. Stopping at the first attempt that ends on its own gives 42.9\%, and the eight attempts give 52.2\%. \emph{Delethink.} After each chunk of $C$ tokens the context is reset to the question, the first 100 tokens of the first chunk, and the last $M = C/2$ tokens of the previous chunk. The solver is the untrained base model; the Markovian Thinker trains it with RL, and we do not. We use $C = 8{,}192$ for the 4B solver and $C = 4{,}096$ for R1-Distill-1.5B, with the sampling settings and prompt the authors provide for that model.

\paragraph{Scoring.}
The answer is the last \texttt{\textbackslash boxed\{\}} after \texttt{</think>}, or the content of \texttt{<answer>}\ldots\texttt{</answer>} if there is none. AIME and BeyondAIME answers are compared as integers, HMMT with \texttt{math-verify}, and Reasoning Gym with its own verifier (score at least 1), parsing grid puzzles into grids compared cell by cell. Accuracies are avg@$k$: the mean over the $k$ runs of a problem, averaged over problems with equal weight.

\paragraph{Scaling with forced answers.}
\cref{fig:scaling_math_forced,fig:scaling_rg_forced} show the runs of \cref{fig:scaling_math,fig:scaling_rg} with forced answers. A run that uses up its budget is forced to answer, and this answer is scored. At 128k, Trained exceeds Single by 7.7 percentage points on mathematics with the 4B solver and by 4.1 with the R1 solver. On Reasoning Gym the gains are 11.0 points on Zebra, 3.9 on Sokoban, and 2.0 on Futoshiki.

\begin{figure*}[!htbp]
    \centering
    \includegraphics[width=0.73\textwidth]{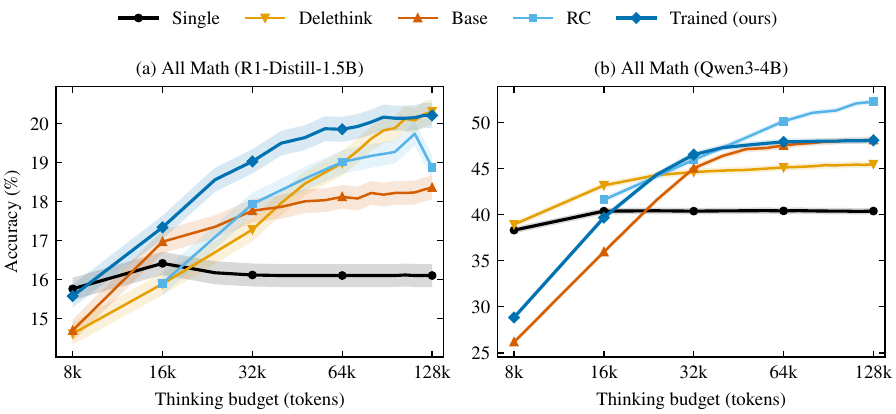}
    \caption{Math scaling with forced answers. The runs are those of \cref{fig:scaling_math}. A run that uses up its budget is forced to answer, and this answer is scored.}
    \label{fig:scaling_math_forced}
\end{figure*}

\begin{figure*}[!htbp]
    \centering
    \includegraphics[width=\textwidth]{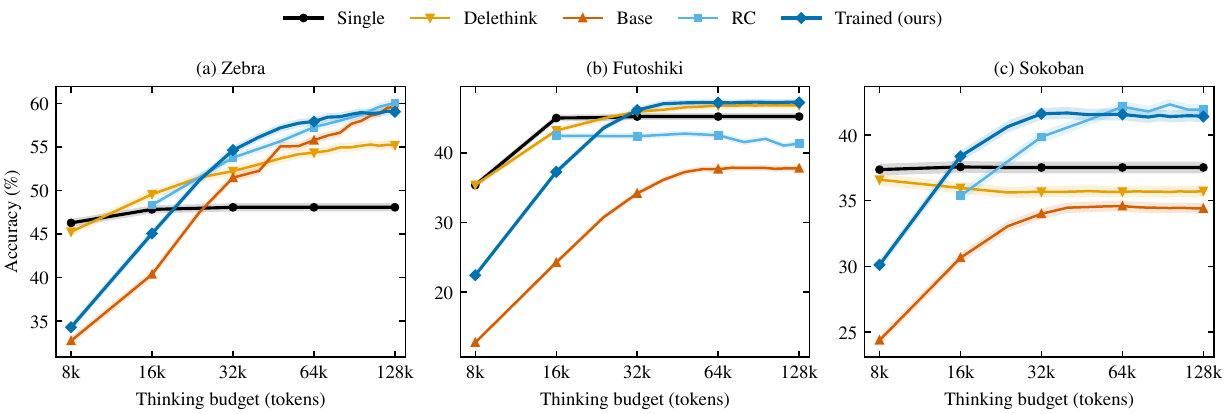}
    \caption{RG scaling with forced answers. The runs are those of \cref{fig:scaling_rg}. A run that uses up its budget is forced to answer, and this answer is scored.}
    \label{fig:scaling_rg_forced}
\end{figure*}

\section{Summary Evaluation Metrics}\label{app:summary_metrics}

{We compare Base and Trained on the same reasoning contexts to measure the effect of training. The four metrics reported in \cref{sec:summary_behavior} (\cref{tab:summary_mechanical}) check what the summaries retain, whether they agree with the source, and how the solver continues from them.} We use 512 reasoning contexts for each setting. Each context is the reasoning of a run before its first cut.

Let $H$ be the source reasoning and $S$ the summary, excluding the retained tail. We extract a set of items $I(X)$ from each text $X$ using fixed rules for its task. These items include numerical facts, cell assignments, coordinates, and moves. We apply the same rules to Base and Trained and compare $I(S)$ with $I(H)$. 

\paragraph{Source-item recall.}
We measure recall as the fraction of extracted source items retained in the summary:
\[
    \operatorname{Recall}(S,H)
    = \frac{|I(H)\cap I(S)|}{|I(H)|}.
\]
Higher recall means that the summary retains more of the items our rules extract.

\paragraph{Bad-item rate.}
The extraction rules flag summary items that are absent from the source or conflict with a value recorded there. These items form $I_{\mathrm{bad}}(S;H)\subseteq I(S)$. We define
\[
    \operatorname{Bad}(S,H)
    = \frac{|I_{\mathrm{bad}}(S;H)|}{|I(S)|}.
\]
A lower rate means better agreement with the source. 

\paragraph{Cap-hit rate.}
We count a cap hit when the token limit cuts a summary off before it finishes. For a batch $\mathcal E$ of generated summaries, the rate is:
\[
    \operatorname{CapHit}(\mathcal E)
    = \frac{1}{|\mathcal E|}\sum_{i\in\mathcal E}
    \mathbf{1}\{\text{summary }i\text{ was stopped by the token limit}\}.
\]
The summary limits are 128 tokens for Base128 and 512 tokens for Base and Trained.

\paragraph{Continuation overlap.}
For the same input, let $c$ be a continuation generated from the
summary-conditioned context and let $\mathcal R$ be the full-context
reference continuations. We measure the fraction of word 4-grams in $c$
that also appear in $\mathcal R$:
\[
    \operatorname{Overlap}(c,\mathcal R)
    = \frac{\#\text{matching word 4-grams in }c}
           {\#\text{word 4-grams in }c}.
\]
For each input and context condition, we average the overlap scores of eight sampled continuations, each up to 1,024 tokens long. Full uses the uncompressed reasoning history. The score compares wording with the reference continuations and does not assess the final answer's correctness.

\section{{Additional Details on Computational Costs}}\label{app:compute}

\subsection{Inference Cost Measurement}\label{app:inference_compute}

The costs in \cref{fig:inference_efficiency} are obtained in two steps. We first measure how the cost of a run grows with the number of solver tokens it has written. On one H100 GPU with 80 GB of memory, we run each method on two mathematics problems with the Qwen3-4B-Instruct solver, one sequence at a time. The solver is not allowed to stop, so every run writes the full 128k budget. For each solver or summarizer call we record its wall-clock time, the peak KV-cache usage of the engine, sampled every 0.2 seconds, and the FLOPs reported by vLLM. Model loading is excluded. The cost at a budget is the sum over the calls made before that budget plus one forced answer, and the memory is the maximum over these calls. We read these costs every 4k tokens and average them over the two problems.

We then combine these cost curves with the 128k evaluation runs, in which every method has 7{,}040 runs (220 problems with 32 runs each). A run that stops after $t$ solver tokens is assigned the cost of its method at $t$. A run that has not stopped at the budget is assigned the cost at the budget, which includes the forced answer. \cref{fig:inference_efficiency} reports the average over the 7{,}040 runs. On ten problems we compared this estimate with a direct measurement, and the two differ by less than 1\% for Single, Trained and Delethink. The cost of RC is approximate. Each RC attempt starts from an empty trace, so we take its cost from the Single curve at the length of the attempt and add the measured cost of the calls that write the cache.

A run that uses the whole 128k budget takes 17.1 minutes, 18.0 GB of KV cache and 6.2 PFLOPs with Single. With TruSuCo it takes 10.9 minutes, 1.0 GB and 2.8 PFLOPs.

\FloatBarrier

\subsection{Training Costs}\label{app:training_costs}

Table~\ref{tab:training_costs} reports training time and GPU resource usage for all five configurations through 80 optimization steps. All GPUs are H100s. GPU allocations include both summarizer training and scoring by the frozen solver.

\begin{table*}[!htbp]
    \centering
    \caption{Training time and GPU resource usage through 80 optimization steps. Step timing reports GPU time computed from measured optimization-step durations. Allocation reports actual allocated GPU time amortized over 80 optimization steps. Values are rounded. R1-1.5B denotes DeepSeek-R1-Distill-Qwen-1.5B.}
    \label{tab:training_costs}
    \small
    \setlength{\tabcolsep}{4pt}
    \renewcommand{\arraystretch}{1.15}
    \begin{tabular*}{\textwidth}{@{\extracolsep{\fill}}llccrrr@{}}
        \toprule
        Experiment
        & \shortstack{Summarizer $\to$\\frozen solver}
        & \shortstack{GPUs\\train + score}
        & \shortstack{Mean step\\time (s)}
        & \shortstack{GPU-s/step\\Step timing}
        & \shortstack{GPU-s/step\\Allocation}
        & \shortstack{Total\\GPU-hours} \\
        \midrule
        Math & 4B $\to$ 4B & $2+1$ & 238 & 715 & 809 & 18.0 \\
        Zebra & 4B $\to$ 4B & $2+1$ & 249 & 746 & 757 & 16.8 \\
        Sokoban & 4B $\to$ 4B & $2+1$ & 215 & 644 & 655 & 14.6 \\
        Futoshiki & 4B $\to$ 4B & $2+1$ & 217 & 651 & 665 & 14.8 \\
        \midrule
        Math (R1) & 1.7B $\to$ R1-1.5B & $1+1$ & 167 & 334 & 362 & 8.1 \\
        \bottomrule
    \end{tabular*}
\end{table*}

\section{Examples of Base and Trained Summaries}\label{app:examples}

This appendix shows summaries written by the untrained and the trained summarizer from identical reasoning. In each example the two runs have the same problem, the same sample and an identical first chunk, so the first summary is the only difference between them. The examples are chosen among problems on which Trained is clearly more accurate than Base, to show what the difference between the summaries looks like. Texts are shown as generated, and \exomit{} marks text that we omit.

\subsection{Mathematics}\label{app:maths_example}

HMMT February 2025, R1-Distill-Qwen-1.5B solver and Qwen3-1.7B summarizer, 128k budget with sink8. The two runs share the first reasoning chunk, so their first summaries are written from identical input. The correct answer is $9\sqrt{15}$.

\noindent
\begin{exbox}[0.492\textwidth]{Question}
Trapezoid $A B C D$, with $A B \| C D$, has side lengths $A B=11, B C=8, C D=19$, and $D A=4$. Compute the area of the convex quadrilateral whose vertices are the circumcenters of $\triangle A B C, \triangle B C D$, $\triangle C D A$, and $\triangle D A B$.\par
\end{exbox}\hfill
\begin{exbox}[0.492\textwidth]{End of the first chunk (excerpts)}
\exomit{}\par
\exgap
Alright, last triangle: DAB.\par
\exgap
Points D(-1, sqrt(15)), A(0,0), B(11,0)\par
\exgap
Find circumcenter.\par
\exgap
Compute perpendicular bisectors of two sides.\par
\exgap
Let me take DA and DB.\par
\exgap
\exomit{}\par
\exgap
 So, equation 1: y =\par
\end{exbox}
\par\medskip

\noindent
\begin{exbox}[0.492\textwidth]{\textcolor{black!65}{Base summary 1 (502 tokens, cut at the cap)}}
Okay, so let me recap where I am.\par
\exgap
So, I need to find the circumcenters of the four triangles. For triangle DAB, I found the perpendicular bisectors of DA and DB. The first equation is \$ y = \textbackslash{}frac\{x\}\{\textbackslash{}sqrt\{15\}\} + \textbackslash{}frac\{1\}\{2\textbackslash{}sqrt\{15\}\} + \textbackslash{}frac\{\textbackslash{}sqrt\{15\}\}\{2\} \$. The second equation is \$ y - \textbackslash{}frac\{\textbackslash{}sqrt\{15\}\}\{2\} = \textbackslash{}frac\{12\}\{\textbackslash{}sqrt\{15\}\}(x - 5) \$. Solving these equations, I get the circumcenter of DAB at \$ \textbackslash{}left( \textbackslash{}frac\{17\}\{2\}, \textbackslash{}frac\{11\textbackslash{}sqrt\{15\}\}\{10\} \textbackslash{}right) \$.\par
\exgap
Now, I have the coordinates of all four circumcenters:\par
\exgap
- ABC: \$ \textbackslash{}left( \textbackslash{}frac\{11\}\{2\}, \textbackslash{}frac\{47\textbackslash{}sqrt\{15\}\}\{10\} \textbackslash{}right) \$\par
- BCD: \$ \textbackslash{}left( \textbackslash{}frac\{17\}\{2\}, \textbackslash{}frac\{33\textbackslash{}sqrt\{15\}\}\{10\} \textbackslash{}right) \$\par
- CDA: \$ \textbackslash{}left( \textbackslash{}frac\{17\}\{2\}, \textbackslash{}frac\{11\textbackslash{}sqrt\{15\}\}\{10\} \textbackslash{}right) \$\par
- DAB: \$ \textbackslash{}left( \textbackslash{}frac\{17\}\{2\}, \textbackslash{}frac\{11\textbackslash{}sqrt\{15\}\}\{10\} \textbackslash{}right) \$\par
\exgap
Now, I need to find the area of the quadrilateral formed by these four points. Since three of the points have the same x-coordinate, the quadrilateral is a trapezoid with bases on the vertical line \$ x = \textbackslash{}frac\{17\}\{2\} \$, and the other two points are at \$ x = \textbackslash{}frac\{11\}\{2\} \$ and \$ x = \textbackslash{}frac\{17\}\{2\} \$.\par
\exgap
The vertical distance between the points is \$ \textbackslash{}frac\{47\textbackslash{}sqrt\{15\}\}\{10\} - \textbackslash{}frac\{11\textbackslash{}sqrt\{15\}\}\{10\} = \textbackslash{}frac\{36\textbackslash{}sqrt\{15\}\}\{10\} = \textbackslash{}frac\{18\textbackslash{}sqrt\{15\}\}\{5\} \$.\par
\exgap
The horizontal distance between the two vertical lines is \$ \textbackslash{}frac\{17\}\{2\} - \textbackslash{}frac\{11\}\{2\} = 3 \$.\par
\end{exbox}\hfill
\begin{exbox}[0.492\textwidth]{\textcolor{Teal}{Trained summary 1 (364 tokens, finished on its own)}}
Okay, so let me recap where I am.\par
\exgap
So, I started by assigning coordinates to the trapezoid ABCD, placing A at (0, 0) and B at (11, 0), then deduced coordinates for D and C based on the given side lengths and the fact that AB is parallel to CD. I solved for x and h, finding x = -1 and h = sqrt(15). This gave me the coordinates: A(0, 0), B(11, 0), D(-1, sqrt(15)), and C(18, sqrt(15)).\par
\exgap
Next, I found the circumcenters of triangles ABC, BCD, CDA, and DAB by computing perpendicular bisectors of their sides. For triangle ABC, I found the perpendicular bisector of AB and BC, and their intersection gave the circumcenter at (11/2, 47\exsym{\surd}15/10). For triangle BCD, I calculated the perpendicular bisectors of BC and BD, and their intersection gave (17/2, 33\exsym{\surd}15/10). For triangle CDA, I used the perpendicular bisector of CD and CA, leading to (17/2, 11\exsym{\surd}15/10). For triangle DAB, I computed the perpendicular bisectors of DA and DB, but the calculation was incomplete due to the complexity of the equations.\par
\exgap
I have finished computing the coordinates of the circumcenters, but the final step of finding the area of the quadrilateral formed by these points is still pending. I have not yet completed the calculations for the last triangle's circumcenter, and the final area computation remains to be done.\par
\exgap
My last line: Simplify: y = (x)/(sqrt(15)) +\par
\end{exbox}
\par\medskip

{Base drops the vertex coordinates and gives DAB the same circumcenter as CDA, $(17/2,11\sqrt{15}/10)$, a result the solver had not reached. Trained keeps the coordinates and the completed results, and records that DAB is unfinished.}

\noindent
\begin{exbox}[0.492\textwidth]{\textcolor{black!65}{Base: the solver\textquotesingle{}s continuation (start)}}
8 sqrt(15)/15.\par
\exgap
Then, the area would be the base times the height: 3 * (18 sqrt(15)/5) = 54 sqrt(15)/5.\par
\exgap
Wait, but hold on, is the quadrilateral a trapezoid? The two points on the vertical line have the same y-coordinate? Let me check.\par
\exgap
 Wait, ABC's circumcenter is at \$ \textbackslash{}left( \textbackslash{}frac\{11\}\{2\}, \textbackslash{}frac\{47\textbackslash{}sqrt\{15\}\}\{10\} \textbackslash{}right) \$, and DAB's is at \$ \textbackslash{}left( \textbackslash{}frac\{17\}\{2\}, \textbackslash{}frac\{11\textbackslash{}sqrt\{15\}\}\{10\} \textbackslash{}right) \$. So, their y-coordinates are different. Similarly, CDA is at \$ \textbackslash{}left( \textbackslash{}frac\{17\}\{2\}, \textbackslash{}frac\{11\textbackslash{}sqrt\{15\}\}\{10\} \textbackslash{}right) \$ and BCD is at \$ \textbackslash{}left( \textbackslash{}frac\{17\}\{2\}, \textbackslash{}frac\{33\textbackslash{}sqrt\{15\}\}\{10\} \textbackslash{}right) \$.\par
\exgap
So, the quadrilateral has four vertices: two at \$ \textbackslash{}left( \textbackslash{}frac\{\par
\exomit{}\par
\end{exbox}\hfill
\begin{exbox}[0.492\textwidth]{\textcolor{Teal}{Trained: the solver\textquotesingle{}s continuation (excerpts)}}
x/sqrt(15) + 8 sqrt(15)/15\par
\exgap
\exomit{}\par
\exgap
Wait, actually, let me double-check my calculations for the last circumcenter. For triangle DAB, I had points D(-1, sqrt(15)), A(0, 0), and B(11, 0). I computed the perpendicular bisector of DA and the perpendicular bisector of DB.\par
\exgap
\exomit{}\par
\exgap
Thus, the area of the convex quadrilateral is \textbackslash{}(\textbackslash{}boxed\{9\textbackslash{}sqrt\{15\}\}\textbackslash{}).\par
\end{exbox}
\par\medskip

With the trained summary, the solver completes DAB at $(11/2,9\sqrt{15}/10)$ and answers $9\sqrt{15}$ after 8{,}064 solver tokens and one summary. With Base, it continues from the unsupported point and answers $277\sqrt{15}/44$ after 22{,}410 tokens and five summaries.

\subsection{Sokoban}
Sokoban at level L0, Qwen3-4B-Instruct solver, 64k budget.

\begin{exbox}[\linewidth]{Puzzle}
You~are~going~to~solve~a~'sokoban'~puzzle.\par
\exgap
*~-~The~player\par
\%~-~The~player~on~a~goal\par
@~-~A~box\par
X~-~A~goal\par
\$~-~A~box~on~a~goal\par
+~-~A~wall\par
-~-~An~empty~position\par
\exgap
Your~solution~must~be~a~string~of~characters,~ex:~LDURRUDL.\par
\exgap
Here~is~your~puzzle:\par
+~+~+~+~+~+\par
+~X~-~-~-~+\par
+~*~X~-~-~+\par
+~-~-~@~-~+\par
+~-~-~@~-~+\par
+~-~-~-~-~+\par
+~+~+~+~+~+\par
\end{exbox}
\par\medskip
\noindent
\begin{exbox}[0.492\textwidth]{\textcolor{black!65}{Base summary 1 (502 tokens, cut at the cap)}}
Okay, so let me recap where I am.\par
\exgap
 So, player at (2,1)\par
Move down to (3,1) \exsym{\rightarrow} D\par
Down to (4,1) \exsym{\rightarrow} D\par
Down to (5,1) \exsym{\rightarrow} D\par
Right to (5,2) \exsym{\rightarrow} R\par
Right to (5,3) \exsym{\rightarrow} R\par
Right to (5,4) \exsym{\rightarrow} R\par
Up to (4,4) \exsym{\rightarrow} U\par
\exgap
Now at (4,4), box at (4,3) \exsym{\rightarrow} push left to (4,2) \exsym{\rightarrow} L\par
\exgap
Now box at (4,2), player at (4,4) \exsym{\rightarrow} can move left to (4,3) \exsym{\rightarrow} L\par
Now at (4,3)\par
\exgap
Push box at (4,2) up to (3,2) \exsym{\rightarrow} U\par
\exgap
Now box at (3,2), player at (4,3) \exsym{\rightarrow} move up to (3,3) \exsym{\rightarrow} U\par
Now at (3,3)\par
\exgap
Push box at (3,2) up to (2,2) \exsym{\rightarrow} U\par
\exgap
Now box at (2,2) \exsym{\rightarrow} goal \exsym{\rightarrow} \$\par
Player at (3,3)\par
\exgap
Now box at (3,3) still there \exsym{\rightarrow} need to move it\par
\exgap
Player at (3,3), box at (3,3)? No -{}-- player just moved to (3,3), box was at (3,2) \exsym{\rightarrow} now moved to (2,2)\par
\exgap
Box at (3,3) is still there \exsym{\rightarrow} yes\par
\exgap
Player at (3,3), box at (3,3) \exsym{\rightarrow} can't push\par
\exgap
Player must move away\par
\exgap
From (3,3) \exsym{\rightarrow} can go to (3,4) \exsym{\rightarrow} R\par
\exgap
Now at (3,4)\par
\exgap
Push box at (3,3) up to (2,3) \exsym{\rightarrow} U\par
\exgap
Now box at (2,3), player at (3,4)\par
\exgap
Push box at (2,3) to (1,3) \exsym{\rightarrow} need player below \exsym{\rightarrow} at (3,3) \exsym{\rightarrow} move to (3,3) \exsym{\rightarrow} L\par
\exgap
Now at (3,3) \exsym{\rightarrow} push up to (2,3) \exsym{\rightarrow} U \exsym{\rightarrow} to (1,3)\par
\exgap
Now box at (1,3), player at (3,3)\par
\exgap
Push to (1,2) \exsym{\rightarrow} need player right \exsym{\rightarrow} at (1,4) or (2,3)\par
\exgap
From (3,3) \exsym{\rightarrow} go\par
\end{exbox}\hfill
\begin{exbox}[0.492\textwidth]{\textcolor{Teal}{Trained summary 1 (387 tokens, finished on its own)}}
Okay, so let me recap where I am.\par
\exgap
So, I assign coordinates (row, col), starting at (0,0).\par
Goals are at (1,1) and (2,2).\par
Boxes are at (3,3) and (4,3).\par
Player starts at (2,1).\par
\exgap
I need both boxes on goals.\par
Boxes can only be pushed if the space behind is free.\par
\exgap
To get a box to (2,2), possible paths: push from (2,3) or (3,2).\par
Currently, no direct access.\par
\exgap
Only possible pushes:\par
- Box at (4,3) can be pushed left to (4,2) if player is at (4,4).\par
- Box at (3,3) can be pushed right to (3,4) if player is at (3,2).\par
\exgap
I decide to move the lower box first.\par
Plan: push (4,3) \exsym{\rightarrow} (4,2) \exsym{\rightarrow} (3,2) \exsym{\rightarrow} (2,2).\par
\exgap
Player must reach (4,4) to push left.\par
Path: (2,1) \exsym{\rightarrow} D \exsym{\rightarrow} (3,1) \exsym{\rightarrow} D \exsym{\rightarrow} (4,1) \exsym{\rightarrow} D \exsym{\rightarrow} (5,1) \exsym{\rightarrow} R \exsym{\rightarrow} (5,2) \exsym{\rightarrow} R \exsym{\rightarrow} (5,3) \exsym{\rightarrow} R \exsym{\rightarrow} (5,4) \exsym{\rightarrow} U \exsym{\rightarrow} (4,4).\par
\exgap
Now player at (4,4), adjacent to (4,3).\par
Can push box left to (4,2).\par
\exgap
Result: box moves from (4,3) to (4,2).\par
Now box at (4,2).\par
\exgap
I will need to push it up later.\par
\exgap
Result: one box now at (4,2), player at (4,4).\par
\exgap
\#\#\# MY LAST LINE\par
Yes -{}-- pushing left from (4,4) would move\par
\end{exbox}
\par\medskip
 The base summary replays the most recent moves until it reaches the 512-token cap and never states the board. The trained summary states the coordinate convention, the goals, the boxes, the plan and the current position, and stops on its own. With the trained summarizer the run solves the puzzle after 4 summaries and 19{,}098 solver tokens. With the base summarizer the run writes 16 summaries, uses the whole 65{,}536-token budget, and its forced answer is wrong.

\subsection{Futoshiki}
Futoshiki at level L0 ($5\times5$), Qwen3-4B-Instruct solver, 64k budget.

\begin{exbox}[\linewidth]{Puzzle}
Solve~the~following~5x5~Futoshiki~puzzle:\par
\exgap
4~>~2~~~\_~~~\_~>~\_\par
~~~~~~~~~~~~~~~~\exsym{\wedge}\par
3~~~\_~>~\_~~~\_~~~\_\par
\exgap
\_~>~\_~~~\_~~~1~~~\_\par
\exgap
\_~~~\_~~~\_~~~2~~~\_\par
~~~~~~~~\exsym{\vee}\par
\_~~~\_~~~3~~~\_~~~5\par
\end{exbox}
\par\medskip
\noindent
\begin{exbox}[0.492\textwidth]{\textcolor{black!65}{Base summary 1 (502 tokens, cut at the cap)}}
Okay, so let me recap where I am.\par
\exgap
So, (3,3) and (3,5) \exsym{\in} \{2,4\}\par
\exgap
Row 3: (3,3), (3,5) = 2 and 4 in some order\par
\exgap
Column 3: missing 1,2,4 \exsym{\rightarrow} (2,3), (3,3), (4,3) \exsym{\in} \{1,2,4\}\par
\exgap
(2,3) \exsym{\in} \{1,2\} from row 2 constraint\par
\exgap
So (3,3) \exsym{\in} \{2,4\}, so possible\par
\exgap
Now row 2: (2,3) \exsym{\in} \{1,2\}, (2,5) \exsym{\in} \{1,2\} -{}-- different values\par
\exgap
Column 5:\par
\exgap
(1,5)=1\par
(2,5)=?\par
(3,5)=?\par
(4,5)=?\par
(5,5)=5\par
\exgap
So column 5: 1, ?, ?, ?, 5 \exsym{\rightarrow} missing: 2,3,4\par
\exgap
So (2,5), (3,5), (4,5) \exsym{\in} \{2,3,4\}\par
\exgap
(2,5) \exsym{\in} \{1,2\} \exsym{\rightarrow} intersection: only 2\par
\exgap
Therefore: (2,5)=2\par
\exgap
Then row 2: (2,3)=1 (since only 1,2 left, and (2,5)=2)\par
\exgap
Now row 2: [3,4,1,5,2]\par
\exgap
Now column 3: (2,3)=1 \exsym{\rightarrow} so missing: 2,4 \exsym{\rightarrow} in (3,3), (4,3)\par
\exgap
Row 3: (3,3) \exsym{\in} \{2,4\} \exsym{\rightarrow} good\par
\exgap
Now column 5: (2,5)=2 \exsym{\rightarrow} so values: 1,2,?,?,5 \exsym{\rightarrow} missing: 3,4 \exsym{\rightarrow} in (3,5), (4,5)\par
\exgap
Row 3: (3,5) \exsym{\in} \{2,4\} \exsym{\rightarrow} but 2 already in column 5 \exsym{\rightarrow} so (3,5) \exsym{\neq} 2 \exsym{\rightarrow} must be 4\par
\exgap
Therefore: (3,5)=4\par
\exgap
Then row 3: (3,3)=2 (only left)\par
\exgap
Now row 3: [5,3,2,1,4]\par
\exgap
Now column 3: (3,3)=2 \exsym{\rightarrow} so missing: 4 \exsym{\rightarrow} so (4,3)=\par
\end{exbox}\hfill
\begin{exbox}[0.492\textwidth]{\textcolor{Teal}{Trained summary 1 (480 tokens, finished on its own)}}
Okay, so let me recap where I am.\par
\exgap
So,\par
I start with the puzzle and label rows and columns.\par
\exgap
Given:\par
- (1,1)=4, (1,2)=2, (1,4) > (1,5), and (1,4) < (2,4)\par
- (2,1)=3, (2,2) > (2,3)\par
- (3,4)=1, (4,4)=2, (5,3)=3, (5,5)=5\par
- Vertical: (1,4) < (2,4), (4,2) > (5,2)\par
\exgap
Row 1: missing 1,3,5. (1,4) \exsym{\in} \{3,5\}, but (1,4) < (2,4) \exsym{\rightarrow} so (1,4)=3, (1,5)=1, (1,3)=5.\par
Row 1: [4,2,5,3,1]\par
\exgap
Column 4: now (1,4)=3, (3,4)=1, (4,4)=2 \exsym{\rightarrow} missing 4,5 \exsym{\rightarrow} (2,4)=5, (5,4)=4\par
\exgap
Row 5: (5,1)=2, (5,2)=1, (5,3)=3, (5,4)=4, (5,5)=5 \exsym{\rightarrow} [2,1,3,4,5]\par
\exgap
Column 1: (3,1)=5, (4,1)=1, (5,1)=2\par
\exgap
Row 2: (2,2)=4, (2,4)=5 \exsym{\rightarrow} missing 1,2 \exsym{\rightarrow} (2,3)\exsym{\in}\{1,2\}, (2,5)=other\par
\exgap
Row 3: (3,1)=5, (3,2)=3 \exsym{\rightarrow} missing 2,4 \exsym{\rightarrow} (3,3),(3,5)\exsym{\in}\{2,4\}\par
\exgap
Column 2: (3,2)=3, (4,2)=5\par
\exgap
Column 3: missing 1,2,4 \exsym{\rightarrow} (2,3)\exsym{\in}\{1,2\}, others to fill\par
\exgap
All constraints used.\par
\exgap
\#\#\# MY LAST LINE\par
Missing: 1,2,4 \exsym{\rightarrow} so (2,\par
\end{exbox}
\par\medskip
 The base summary copies a chain of local deductions and is cut at the cap. The trained summary lists the givens, the rows and columns that are already settled, and the remaining candidates. With the trained summarizer the run solves the puzzle after 1 summary and 6{,}894 solver tokens. With the base summarizer the run writes 16 summaries, uses the whole 65{,}536-token budget, and its forced answer is wrong.

\subsection{Zebra}
Zebra at level L2 (7 houses, 6 attributes), Qwen3-4B-Instruct solver, 128k budget with sink8. The correct answer is Emily.

\noindent
\begin{exbox}[0.492\textwidth]{Question (start)}
This is a logic puzzle. There are 7 houses (numbered 1 on the left, 7 on the right), from the perspective of someone standing across the street from them. Each has a different person in them. They have different characteristics:\par
 - Each person has a unique name: david, bob, arnold, carol, eric, emily, alice\par
 - Each person has a favorite color: brown, yellow, green, purple, blue, white, red\par
 - Everyone has a different favorite cigar: blue master, yellow monster, dunhill, blends, pall mall, prince, red eye\par
 - Each mother is accompanied by their child: billy, meredith, timothy, alice, samantha, fred, bella\par
 - Everyone has a favorite smoothie: dragonfruit, lilac, lemon, life, darkness, butterscotch, desert\par
 - The people keep different animals: cat, bird, horse, rabbit, pig, fish, dog\par
\exgap
1. David is directly left of the rabbit owner.\par
2. The Desert smoothie lover is the person whose favorite col\par
\exomit{}\par
\end{exbox}\hfill
\begin{exbox}[0.492\textwidth]{End of the first chunk, identical in both runs}
Life\par
\exgap
Now, red=2 \exsym{\rightarrow} house 2 \exsym{\rightarrow} color red \exsym{\rightarrow} smoothie: Desert\par
\exgap
House 2: Alice \exsym{\rightarrow} so Alice \exsym{\rightarrow} color: red\par
\exgap
Now, clue 7: Eric \exsym{\rightarrow} yellow\par
\exgap
So someone has yellow.\par
\exgap
Now, clue 1: David is directly left of rabbit owner.\par
\exgap
David \exsym{\rightarrow} child: Alice\par
\exgap
We don't know where David is.\par
\exgap
Names so far:\par
- H2: Alice\par
- H3: Arnold\par
- H4: Bob\par
\exgap
Remaining: David, Carol, Eric, Emily \exsym{\rightarrow} for H1, H5, H6, H7\par
\exgap
Now, clue 19: child Fred \exsym{\rightarrow} directly left of Blue Master smoker\par
\exgap
And child Fred \exsym{\rightarrow} animal: pig\par
\exgap
So pig owner \exsym{\rightarrow} directly left of Blue Master smoker\par
\exgap
Animals:\par
- H2: Alice \exsym{\rightarrow} dog\par
- H4: Bob \exsym{\rightarrow} fish\par
- pig \exsym{\rightarrow} in some house, left of Blue Master\par
\exgap
Animals left: cat, rabbit, horse, pig, bird (Emily has bird)\par
\exgap
Emily \exsym{\rightarrow} bird\par
\exgap
So bird is somewhere.\par
\end{exbox}
\par\medskip
\noindent
\begin{exbox}[0.492\textwidth]{\textcolor{black!65}{Base summary 1 (502 tokens, cut at the cap)}}
Okay, so let me recap where I am.\par
\exgap
So, Alice in X, Arnold in X+1, Bob in X+2\par
Bob \exsym{\neq} house 3 \exsym{\rightarrow} X \exsym{\neq} 1\par
X \exsym{\in} \{2,3,4,5\}\par
\exgap
Try X=2:\par
H2: Alice, H3: Arnold, H4: Bob\par
H3 cigar: Prince \exsym{\rightarrow} Arnold smokes Prince\par
Bob: color green, cigar Yellow Monster, smoothie Dragonfruit, animal fish\par
Arnold: smoothie Butterscotch, child Bella\par
\exgap
Clue 17: brown left of white\par
House 4 is green \exsym{\rightarrow} so white \exsym{\neq} 4\par
(3,4): white=4 \exsym{\rightarrow} invalid\par
(4,5): white=5 \exsym{\rightarrow} brown=4 \exsym{\rightarrow} green \exsym{\rightarrow} invalid\par
So only (1,2), (2,3), (5,6), (6,7)\par
(2,3): white=3 \exsym{\rightarrow} brown=2\par
(1,2): white=2 \exsym{\rightarrow} brown=1\par
(5,6), (6,7)\par
\exgap
Clue 25: blends left of red\par
red \exsym{\neq} 3,4 \exsym{\rightarrow} red \exsym{\in} \{2,5,6,7\}\par
red=5 \exsym{\rightarrow} blends=4 \exsym{\rightarrow} but H4 cigar=Yellow Monster \exsym{\rightarrow} invalid\par
red \exsym{\neq} 5\par
red \exsym{\in} \{2,6,7\}\par
\exgap
red=2 \exsym{\rightarrow} H2 color red \exsym{\rightarrow} smoothie Desert\par
H2 is Alice \exsym{\rightarrow} Alice has red color\par
\exgap
brown left of white \exsym{\rightarrow} brown \exsym{\in} \{1,2,5,6\}\par
but brown \exsym{\neq} 2 if red=2 \exsym{\rightarrow} so brown \exsym{\in} \{1,5,6\}\par
brown left of white \exsym{\rightarrow} so (1,2): white=2 \exsym{\rightarrow} but red=2 \exsym{\rightarrow} conflict\par
(2,3): brown=2 \exsym{\rightarrow} red=2 \exsym{\rightarrow} conflict\par
(5,6): white=6 \exsym{\rightarrow} brown=5\par
(6,7): white=7 \exsym{\rightarrow} brown=6\par
\exgap
So brown in 5 or 6 \exsym{\rightarrow} smoothie Life in 5 or 6\par
\exgap
Smoothies:\par
H3: Butterscotch, H4: Dragonfruit\par
red=2 \exsym{\rightarrow} H2: Desert\par
So Life in 5 or 6 \exsym{\rightarrow} ok\par
\exgap
Now clue 1: David left of rabbit owner\par
David has child Alice\par
\exgap
Names: H2: Alice, H3: Arnold, H4: Bob \exsym{\rightarrow} left: David, Carol, Eric, Emily\par
\exgap
H\par
\end{exbox}\hfill
\begin{exbox}[0.492\textwidth]{\textcolor{Teal}{Trained summary 1 (390 tokens, finished on its own)}}
Okay, so let me recap where I am.\par
\exgap
So,\par
House 3 \exsym{\rightarrow} cigar: Prince\par
\exgap
Bob \exsym{\rightarrow} color: green, smoothie: Dragonfruit, animal: fish, cigar: Yellow Monster\par
\exgap
Alice \exsym{\rightarrow} cigar: Dunhill, animal: dog, child: Samantha\par
\exgap
Arnold \exsym{\rightarrow} child: Bella, smoothie: Butterscotch\par
\exgap
Alice is directly left of Arnold and Butterscotch drinker \exsym{\rightarrow} same house\par
\exgap
Thus: Alice \exsym{\rightarrow} left of \exsym{\rightarrow} Arnold \exsym{\rightarrow} drinks Butterscotch\par
\exgap
Bob is adjacent to Arnold (clue 13) \exsym{\rightarrow} Bob in Alice + 2\par
\exgap
Bob \exsym{\neq} house 3 (cigar conflict) \exsym{\rightarrow} eliminate X=1\par
\exgap
Possible: X=2,3,4,5\par
\exgap
Try X=2:\par
- H2: Alice\par
- H3: Arnold\par
- H4: Bob\par
\exgap
House 3: Prince \exsym{\rightarrow} ok\par
\exgap
Bob: green, fish, Dragonfruit, Yellow Monster\par
\exgap
Arnold: Butterscotch, child: Bella\par
\exgap
Clue 17: brown left of white\par
Clue 25: blends left of red\par
\exgap
Red \exsym{\rightarrow} smoothie: Desert\par
Brown \exsym{\rightarrow} smoothie: Life\par
\exgap
House 4: green \exsym{\rightarrow} not brown/white\par
\exgap
Red \exsym{\in} \{2,6,7\} (not 3,4)\par
blends in red-1\par
\exgap
Try red=2: H2 color=red \exsym{\rightarrow} smoothie=Desert\par
Then brown \exsym{\in} \{5,6\} (from brown left of white)\par
\exgap
Smoothies:\par
H3: Butterscotch\par
H4: Dragonfruit\par
H2: Desert\par
\exgap
Brown \exsym{\rightarrow} Life \exsym{\rightarrow} in 5 or 6\par
\exgap
Eric \exsym{\rightarrow} yellow\par
Emily \exsym{\rightarrow} bird\par
\exgap
David \exsym{\rightarrow} child: Alice, left of rabbit owner\par
\exgap
Names: Alice (H2), Arnold (H3), Bob (H4) \exsym{\rightarrow} rest: David, Carol, Eric, Emily\par
\exgap
\#\#\# MY LAST LINE\par
-\par
\end{exbox}
\par\medskip
 With the trained summarizer the run writes 3 summaries of 390, 396 and 405 tokens, all finished on their own, and answers correctly after 13{,}397 solver tokens. Each summary is a table of assignments per house followed by the open case, and the table grows from one summary to the next. With the base summarizer the run writes 32 summaries, nearly all cut at the cap, and uses the whole 131{,}072-token budget. Its summaries copy the derivation, and the solver starts the placement again with a different hypothesis each time. Its forced answer is wrong.

\noindent
\begin{minipage}[t]{0.492\textwidth}
\begin{exbox}[\linewidth]{\textcolor{black!65}{Base summary 13 (start)}}
Okay, so let me recap where I am.\par
\exgap
So, Alice=4 \exsym{\rightarrow} Arnold=5 \exsym{\rightarrow} Bob=6\par
\exgap
House 4: Alice \exsym{\rightarrow} Dunhill, child Samantha\par
House 5: Arnold \exsym{\rightarrow} child Bella, smoothie Butterscotch, cigar Prince\par
House 6: Bob \exsym{\rightarrow} green, fish, Dragonfruit, Yellow Monster\par
\exgap
Clue 1: David directly left of rabbit \exsym{\rightarrow} David in X, rabbit in X+1\par
David \exsym{\in} \{1,2,3,4,5,6\} \exsym{\rightarrow} not 4,5,6 \exsym{\rightarrow} David \exsym{\in} \{1,2,3\}\par
\exgap
Clue 26: David \exsym{\rightarrow}\par
\exomit{}\par
\end{exbox}\par\medskip
\begin{exbox}[\linewidth]{\textcolor{black!65}{Base summary 25 (start)}}
Okay, so let me recap where I am.\par
\exgap
So, Alice directly left of Arnold \exsym{\rightarrow} Alice in X, Arnold in X+1.\par
House 3 has Prince (clue 15).\par
Alice has Dunhill (clue 22).\par
\exgap
 If Alice in 3 \exsym{\rightarrow} house 3 has both Dunhill and Prince \exsym{\rightarrow} conflict.\par
Alice in 2 \exsym{\rightarrow} Arnold in 3 \exsym{\rightarrow} house 3: Arnold, Prince \exsym{\rightarrow} possible.\par
Alice in 1 \exsym{\rightarrow} Arnold in 2 \exsym{\rightarrow} house 2: Arnold \exsym{\rightarrow} possible.\par
Alice in 4 \exsym{\rightarrow} Arnold in 5 \exsym{\rightarrow} h\par
\exomit{}\par
\end{exbox}\par\medskip
\begin{exbox}[\linewidth]{\textcolor{black!65}{Base summary 32, the last one (start)}}
Okay, so let me recap where I am.\par
\exgap
So, red=6 or 7 only possible.\par
 red=6 \exsym{\rightarrow} blends=5 \exsym{\rightarrow} house 5: blends\par
red=7 \exsym{\rightarrow} blends=6 \exsym{\rightarrow} house 6: blends\par
\exgap
Try red=6: house 6: red, Desert, blends=5\par
\exgap
brown left of white: only (1,2) or (2,3)\par
brown in 1 or 2 \exsym{\rightarrow} smoothie Life in that house\par
\exgap
house 2: animal dog (from clue 8)\par
\exgap
fish in 4 \exsym{\rightarrow} Bob=4 \exsym{\rightarrow} good\par
\exgap
David in 1,5,6\par
\exgap
David=1 \exsym{\rightarrow} rabbit=2\par
\exomit{}\par
\end{exbox}
\end{minipage}\hfill
\begin{exbox}[0.492\textwidth]{\textcolor{Teal}{Trained summary 3, the last one}}
Okay, so let me recap where I am.\par
\exgap
So,\par
H2: color=red (Desert)\par
H4: color=green (Dragonfruit)\par
H3: cigar=Prince\par
\exgap
Child Fred = H5 \exsym{\rightarrow} animal=pig \exsym{\rightarrow} Blue Master = H6\par
David = H6 \exsym{\rightarrow} child=Alice\par
\exgap
David directly left of rabbit \exsym{\rightarrow} rabbit = H7\par
\exgap
Names:\par
H2: Alice\par
H3: Arnold\par
H4: Bob\par
H6: David\par
\exsym{\rightarrow} H1: Emily (only one left with bird)\par
\exgap
Animals:\par
H1: bird\par
H2: dog\par
H4: fish\par
H5: pig\par
H7: rabbit\par
\exsym{\rightarrow} horse \exsym{\in} \{3,6\} \exsym{\rightarrow} purple \exsym{\in} \{3,6\}\par
\exgap
Cigars:\par
H1: blends\par
H2: Dunhill\par
H3: Prince\par
H4: Yellow Monster\par
H6: Blue Master\par
\exsym{\rightarrow} H5, H7: Red Eye, Pall Mall\par
\exgap
Clue 25: blends left of red \exsym{\rightarrow} H1 \exsym{\rightarrow} blends \exsym{\rightarrow} correct\par
\exgap
Clue 17: brown left of white \exsym{\rightarrow} brown \exsym{\in} \{5,6\}\par
\exgap
**Case A: brown=5 \exsym{\rightarrow} white=6**\par
H5: brown \exsym{\rightarrow} smoothie=Life\par
H6: white\par
\exgap
Yellow \exsym{\rightarrow} Eric \exsym{\rightarrow} H7 \exsym{\rightarrow} yellow\par
Purple \exsym{\rightarrow} H3\par
\exgap
Colors:\par
H1: blue\par
All assigned\par
\exgap
Children:\par
H4: Timothy\par
H5: Fred\par
H6: Alice\par
\exsym{\rightarrow} H1, H7: Billy, Meredith\par
\exgap
Billy \exsym{\rightarrow} Pall Mall \exsym{\rightarrow} H7 \exsym{\rightarrow} child=Billy\par
H1 \exsym{\rightarrow} child=Meredith \exsym{\rightarrow} smoothie=Darkness\par
\exgap
Smoothies:\par
H5: Life\par
H1: Darkness\par
\exsym{\rightarrow} H6, H7: Lilac, Lemon\par
\exgap
H7: cigar=Pall Mall \exsym{\rightarrow} not Red Eye \exsym{\rightarrow} no Lilac\par
\exsym{\rightarrow} H7: L\par
\exomit{}\par
\end{exbox}
\par\medskip
\noindent
\begin{exbox}[0.492\textwidth]{\textcolor{black!65}{Base: end of the run}}
After a very long and complex deduction, the person in house 1 is **Carol**.\par
\exgap
\textbackslash{}boxed\{Carol\}\par
\end{exbox}\hfill
\begin{exbox}[0.492\textwidth]{\textcolor{Teal}{Trained: end of the run}}
ily must be in H1 \exsym{\rightarrow} only house with bird \exsym{\rightarrow} H1 \exsym{\rightarrow} so H1: Emily\par
\exgap
Names: H1: Emily\par
\exgap
Now all names:\par
\exgap
H1: Emily\par
H2: Alice\par
H3: Arnold\par
H4: Bob\par
H5: Eric\par
H6: David\par
H7: Carol\par
\exgap
Now check clue 26: The person whose child is Alice is David.\par
\exgap
David \exsym{\rightarrow} child=Alice \exsym{\rightarrow} yes \exsym{\rightarrow} H6 \exsym{\rightarrow} child=Alice \exsym{\rightarrow} correct.\par
\exgap
All clues seem to check.\par
\exgap
Now the question is: Who lives in House 1?\par
\exgap
Answer: **Emily**\par
\exgap
\textbackslash{}boxed\{Emily\}\par
\end{exbox}
\par\medskip

\end{document}